\documentclass[letterpaper]{article} 
\usepackage{aaai2027}  

\usepackage[hyphens]{url}  
\usepackage{graphicx} 
\usepackage{natbib}  
\usepackage{caption} 
\usepackage[none]{hyphenat}
\usepackage{amsmath,amssymb,amsfonts}
\usepackage{booktabs}
\usepackage{multirow}
\usepackage{pifont}
\usepackage{xcolor}
\usepackage{colortbl}

\newcommand{\cmark}{\textcolor{green!45!black}{\ding{51}}}
\newcommand{\xmark}{\textcolor{red!60!black}{\ding{55}}}
\newcommand{\ours}{PixelUp}

\title{PixelUp: Zero-Shot Semantic Feature Upsampling for Fine-Grained Vision Tasks}

\author{
    Deepank Singh, Anurag Nihal, Vedhus hoskere
}
\affiliations{
University of Houston
}

\begin{document}

\maketitle

\begin{figure*}
\centering
\includegraphics[width=0.8\linewidth]{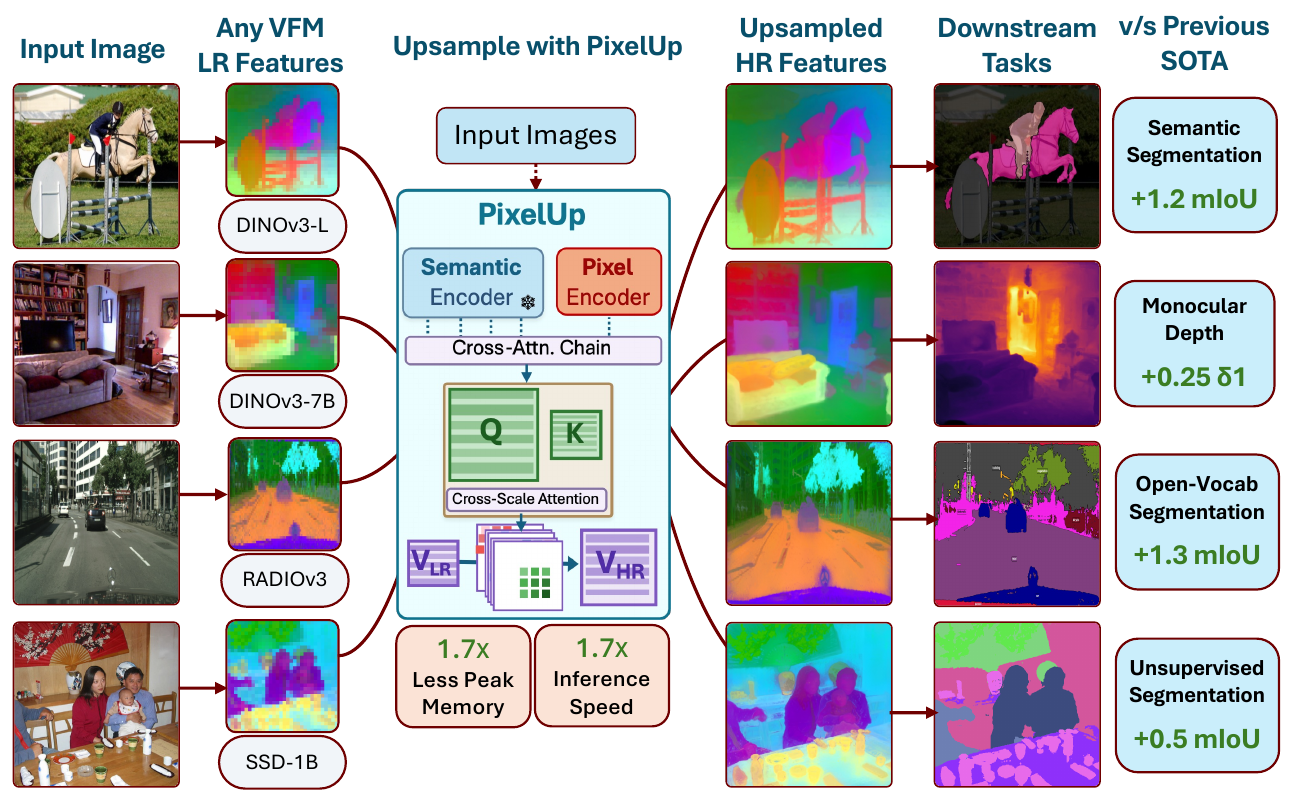}
\caption{\textbf{\ours{} is a zero-shot, VFM-agnostic feature upsampler.} \ours{} lifts the coarse patch tokens of any VFM to dense, pixel-level feature via guidance from the multi-scale features of a frozen Semantic Encoder (DINOv3 ConvNeXt-S). \ours{} outperforms prior VFM-agnostic upsamplers and the VFM-specific JAFAR across diverse dense prediction tasks.}
\label{fig:teaser}
\end{figure*}

\begin{abstract}
Self-supervised Vision Foundation Models (VFMs) have become essential backbones for downstream tasks due to their strong and transferable visual representations. However, their patch-token-level features are often too coarse for dense prediction tasks such as semantic segmentation and depth estimation when accurate fine-grained predictions are required. Feature upsampling methods have been developed to recover pixel-level detail but still face limitations. Learnable upsamplers are often designed for a specific encoders and must be retrained for different encoders. Image-guided methods that use shallow pixel encoders often introduce textural artifacts and lack the semantic guidance needed for accurate downstream predictions. We introduce PixelUp, a zero-shot VFM-agnostic upsampler achieving semantic awareness through a coarse-to-fine chain of windowed cross-attention architecture guided by multi-scale semantic features. We demonstrate that PixelUp outperforms both VFM-specific and VFM-agnostic upsamplers, achieving state-of-the-art performance on dense prediction tasks with an average improvement of +1.2 mIoU on semantic segmentation and +0.25 $\delta_1$, on NYUv2 depth estimation across VFMs. PixelUp further improves training-free open-vocabulary and unsupervised semantic segmentation by an average of +1.3 mIoU and +0.5 mIoU, respectively.

\end{abstract}


\section{Introduction}

Self-supervised vision foundation models (VFMs) provide rich transferable spatial and semantic representations that have been increasingly adopted for dense prediction tasks, including semantic segmentation and monocular depth estimation~\cite{dinov3,radio}. However, many modern VFM backbones encode images as non-overlapping patch-token features where each patch-token summarizes 14-to-16 pixel image region. Although semantically rich, the resulting representations obscure local boundaries and fine-grained details which limits the spatial details available and reduces prediction accuracy in pixel-sensitive tasks such as depth estimation.~\cite{vit,featup,jafar}. A straightforward solution is to increase the input scale and obtain a denser feature grid resolution. However, the self-attention mechanism in these VFMs quadratically increases the compute cost with increase in input resolution. Moreover, the shift in test-time input scale introduces feature artifacts that degrade downstream performance~\cite{ViTAR}. These constraints motivate post-hoc feature upsampling that recovers spatially dense representations while preserving the semantic content encoded by the frozen backbone. Existing methods address this objective through task-specific operators, or learned task-agnostic architectures, but prior approaches do not combine VFM-agnostic transfer with high-resolution semantic guidance. Table~\ref{tab:landscape} summarizes these distinctions. Among the prior existing methods, \ours{} alone combines task-and VFM-agnostic transfer, arbitrary-scale upsampling, and high-resolution semantic guidance.

\begin{table}[t]
\centering
{\small
\setlength{\tabcolsep}{3pt}
\begin{tabular}{@{}lccccc@{}}
\toprule
Method & \textit{TF} & \textit{TA} & \textit{VA} & \textit{AS} & \textit{SG} \\
\midrule
JBU~\cite{jbu}             & \cmark & \cmark & \cmark & \cmark & \xmark \\
FeatUp~\cite{featup}       & \xmark & \cmark & \xmark & \xmark & \xmark \\
LiFT~\cite{lift}           & \xmark & \cmark & \xmark & \xmark & \xmark \\
LoftUp~\cite{loftup}       & \xmark & \cmark & \xmark & \cmark & \xmark \\
FeatSharp~\cite{ranzingerfeatsharp} & \xmark & \cmark & \xmark & \xmark & \cmark \\
JAFAR~\cite{jafar}         & \xmark & \cmark & \xmark & \cmark & \xmark \\
AnyUp~\cite{anyup}         & \xmark & \cmark & \cmark & \cmark & \xmark \\
NAF~\cite{naf}             & \xmark & \cmark & \cmark & \cmark & \xmark \\
\rowcolor{blue!8}
\textbf{\ours{} (ours)}    & \xmark & \cmark & \cmark & \cmark & \cmark \\
\bottomrule
\end{tabular}
}
\caption{\textbf{Overview of feature upsampling methods.}
\textit{TF} denotes training-free operation, \textit{TA} task-agnostic transfer, \textit{VA} VFM-agnostic transfer, \textit{AS} arbitrary-scale upsampling, and \textit{SG} high-resolution guidance conditioned on pretrained semantic features.}
\label{tab:landscape}
\end{table}

Traditional fixed-form interpolation relies on spatial proximity, while JBU and guided filtering use high-resolution RGB similarities but can merge semantically distinct regions with similar appearance~\cite{jbu,guidedfilter}. Task-specific operators such as CARAFE \cite{carafe}, SAPA \cite{sapa}, FADE \cite{fade}, and DySample \cite{dysample} learn reassembly kernels or offsets within encoder-decoder networks and require joint optimization for each task and architecture. Task-agnostic methods such as FeatUp \cite{featup}, LiFT \cite{lift}, and LoftUp \cite{loftup} recover high-resolution features from frozen backbones across tasks, yet each checkpoint remains tied to one encoder feature space~\cite{featup,lift,loftup}.

Recent attention-based upsamplers such as JAFAR \cite{jafar}, AnyUp \cite{anyup}, and NAF \cite{naf} use cross-attention to combine coarse VFM features into dense outputs for upsampling. Across all three methods, a shallow encoder processes the input image to produce high-resolution queries, while the frozen target VFM supplies coarse features as values. Low-resolution keys align with the VFM token grid, and query-key attention determines how the coarse values contribute to each dense output location. In particular, JAFAR forms keys from downsampled image features and modulates them with target-VFM features through Spatial Feature Transform, introducing semantics only at the coarse key level and requiring backbone-specific training. AnyUp applies a feature-agnostic layer to coarse target-VFM features and combines the output with downsampled image features as keys, whereas image pixel features remain the high-resolution queries. NAF forms queries and pooled keys from RGB features produced by $1{\times}1$ pixel and $3{\times}3$ contextual branches, limiting guidance to appearance and local context. Consequently, existing methods either introduce pretrained semantics only through low-resolution keys or exclude them from the guidance pathway, leaving high-resolution queries without explicit semantic conditioning.

We address these limitations with \textbf{\ours{}}, a self-supervised, zero-shot, VFM-agnostic upsampler that uses query-side semantic guidance to upsample coarse patch features from any frozen VFM into dense pixel-level features as shown in Figure \ref{fig:teaser}. Our main contributions are threefold.
\begin{itemize}
    \item We propose \ours{}, a self-supervised, zero-shot, and VFM-agnostic feature upsampler whose key novelty is query-side semantic guidance using multi-scale features from a pretrained semantic encoder. A coarse-to-fine Cross-Attention Chain progressively fuses initial pixel-encoder queries with these multi-scale semantic encoder features, yielding high-resolution, semantically rich queries whose pooled counterparts serve as low-resolution keys for upsampling target-VFM features. 
    
    \item We demonstrate zero-shot transfer from one \ours{} checkpoint across ten frozen VFMs spanning self-supervised, distilled, and diffusion-based models up to 7B parameters. Under matched settings, \ours{} provides $1.6{\times}$-$1.7{\times}$ higher throughput and $1.6{\times}$-$1.9{\times}$ lower peak memory than prior state-of-the-art (SOTA) approaches.
    \item We evaluate \ours{} on supervised semantic segmentation, monocular depth estimation, unsupervised segmentation, and open-vocabulary segmentation. \ours{} surpasses prior SOTAs in different supervised segmentation settings and achieves the highest $\delta_1$ on five of six NYUv2 backbones. As a drop-in upsampler, \ours{} improves DiffCut and RADSeg by $+0.5$ and $+1.3$ mIoU on average.
\end{itemize}

\section{Related Work}

\subsection{VFM-Specific Feature Upsamplers}
VFM-specific upsamplers learn task-agnostic feature reconstruction within the feature space of a particular frozen backbone, requiring retraining for each new backbone. FeatUp minimizes a multi-view consistency loss using either feed-forward stack of learned JBU filters which generalizes across images but tends to over-smooth region interiors, or a per-image implicit network, which produces sharper features but must be re-optimized for every input~\cite{featup}. LiFT is a compact U-Net-style block that doubles the resolution of ViT descriptors in a single forward pass~\cite{lift}. LoftUp uses coordinate-and-RGB queries with coarse VFM features as keys and values, then trains from mask-refined pseudo-features followed by self-distillation~\cite{loftup}. FeatSharp combines FeatUp's JBU output with a tiled mosaic of features extracted by the same backbone and refines their fusion through local attention~\cite{ranzingerfeatsharp}. JAFAR cross-attends at any output scale but binds each checkpoint to one encoder~\cite{jafar}. Thus, although these methods are task-agnostic, their reusable checkpoints remain backbone-specific. In contrast, a single \ours{} checkpoint can upsample features from any VFM.

\subsection{VFM-Agnostic Feature Upsamplers}
A recent line of work upsamples any VFM with a single checkpoint. 
AnyUp canonicalizes dimension-varying target-VFM features and combines them with downsampled RGB features as keys for local-window attention~\cite{anyup}. RGB-derived high-resolution pixel features serve as queries, while unmodified target-VFM features remain values. NAF removes target-VFM features from guidance and derives queries and pooled keys from a RoPE-encoded, dual-branch RGB representation formed by $1{\times}1$ pixel and $3{\times}3$ contextual convolutions~\cite{naf}. Cross-scale neighborhood attention then combines nearby target-VFM values at each output location. Both methods support one-checkpoint VFM transfer, but neither conditions the high-resolution query on pretrained semantic features. \ours{} instead conditions high-resolution queries on multi-scale features from a separate frozen semantic encoder. Target-VFM features remain outside the guidance pathway, preserving one-checkpoint transfer across VFM backbones.

\begin{figure*}
    \centering
    \includegraphics[width=0.95\linewidth]{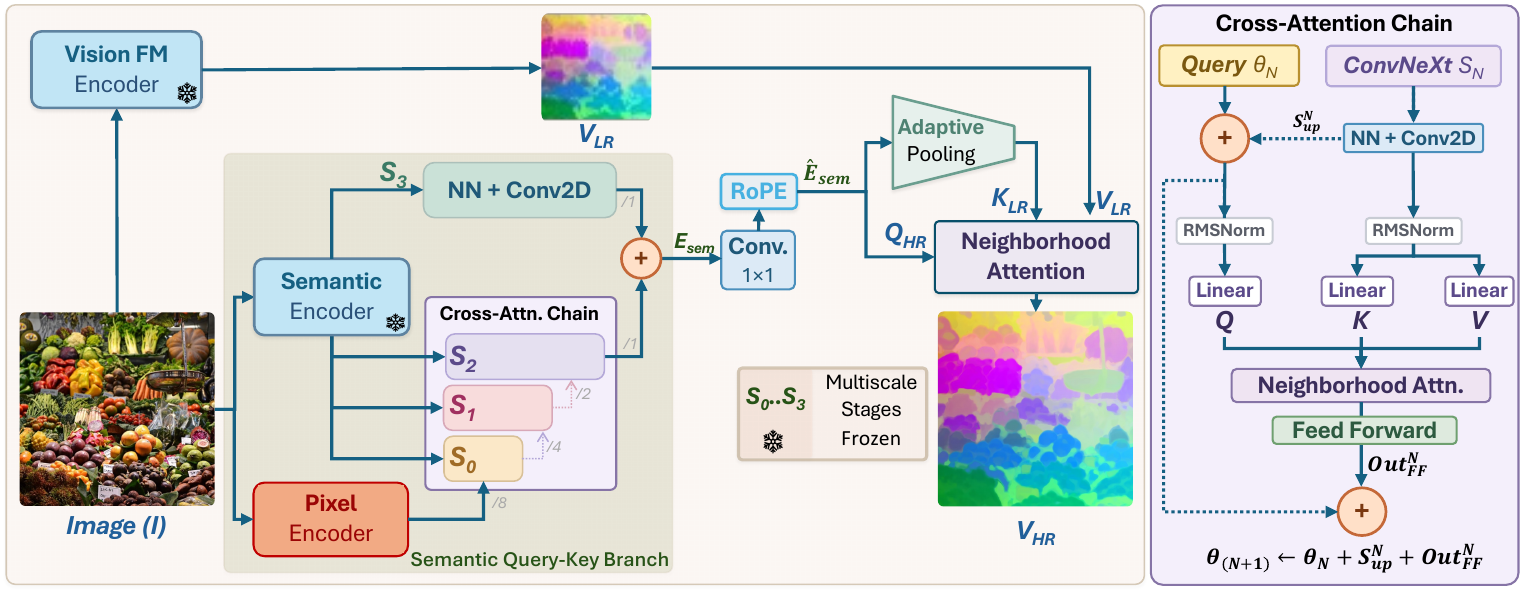}
    \caption{\textbf{\ours{} overview.} \ours{} is a zero-shot, VFM-agnostic upsampler that lifts the coarse patch tokens of any frozen VFM to dense pixel-level features. A Pixel Encoder forms an initial query, and a Cross-Attention Chain refines the query with the multi-scale semantic features of a pre-trained frozen Semantic Encoder into the high-resolution query $Q_{HR}$ and pooled key $K_{LR}$ , which combine the values $V_{LR}$ via neighborhood attention into the dense output $V_{HR}$.}
    \label{fig:architecture}
\end{figure*}

\section{Methodology}


\ours{} is a zero-shot, VFM-agnostic upsampler that reconstructs dense pixel-level features from an input image and coarse features produced by a frozen target VFM. Our architecture separates the preserved target-VFM values from a semantic query-key branch containing a lightweight Pixel Encoder and a frozen Semantic Encoder such as DINOv3 ConvNeXt-S. A coarse-to-fine Cross-Attention Chain integrates multi-scale semantic features into a high-resolution query, while adaptive pooling maps the same representation to the target VFM token grid as the low-resolution key. Local neighborhood attention then combines the target-VFM values according to query-key similarity, allowing one \ours{} checkpoint to transfer across VFM backbones. Figure~\ref{fig:architecture} summarizes the complete forward path.

\subsection{General Formulation}

Given an input image $\mathbf{I}\in\mathbb{R}^{H\times W\times 3}$, a frozen vision foundation model $f$ produces the low-resolution feature map $V_{LR}=f(\mathbf{I})\in\mathbb{R}^{h\times w\times d}$, where $d$ is the channel width of the target Vision FM encoder and $h{=}\frac{H}{p}$, $w{=}\frac{W}{p}$ for patch stride $p$. In our \ours{} architecture, we perform neighborhood attention which requires three attention operands, (i) a high-resolution query, (ii) a low-resolution key, and (iii) low-resolution values to reconstruct a high-resolution dense feature map $V_{HR}\in\mathbb{R}^{H\times W\times d}$. 

The semantic query-key branch derives the query and key from $\mathbf{I}$. The branch first produces a high-resolution semantic encoding $E_{\mathrm{enc}}\in\mathbb{R}^{H\times W\times d_{\mathrm{enc}}}$. A $1{\times}1$ convolution followed by RoPE~\cite{rope} transforms $E_{\mathrm{enc}}$ into the position-encoded semantic representation $\hat{E}_{\mathrm{enc}}$. Finally, both query and key operands are derived from $\hat{E}_{\mathrm{enc}}$. Query $Q_{HR}$ is taken at the full grid, while key $K_{LR}$ is obtained by adaptive average pooling of the same spatial resolution as $V_{LR}$. It is important to note that RoPE is applied before the Query-Key split, so that both the high-resolution query $Q_{HR}$ and the low-resolution key $K_{LR}$ carry consistent positional phases and align across resolutions. 

Next, we compute cross-scale neighborhood attention between $Q_{HR}$ and $K_{LR}$. For each high-resolution position $p$, the query $Q_{HR}^{p}$ attends only to a compact neighborhood $\mathcal{N}_{LR}(p)$ around its corresponding position on the low-resolution grid. As defined in Equation~\ref{eq:pixelup_attn}, the resulting attention scores are normalized by $Z(p)$ as shown in Equation \ref{eq:zp} and then used to aggregate the low-resolution values $V_{LR}^{q}$ within this neighborhood to obtain the upsampled feature $V_{HR}^{p}$. The query-key branch determines only the scalar spatial weights, while the target VFM features $V_{LR}$ remain independent. This separation allows the same attention mechanism to operate across VFMs with different feature dimensions, which enables zero-shot transfer and makes \ours{} a VFM-agnostic feature upsampler.
\begin{equation}
\begin{aligned}
V_{HR}^{p}
&=
\frac{1}{Z(p)}
\sum_{q\in\mathcal{N}_{LR}(p)}
\exp\!\left[
\frac{
\left\langle Q_{HR}^{p}, K_{LR}^{q}\right\rangle
}{
\sqrt{d_{enc}}
}
\right]
V_{LR}^{q}
\end{aligned}
\label{eq:pixelup_attn}
\end{equation}
\begin{equation}\label{eq:zp}
    Z(p) = \sum_{q'\in\mathcal{N}_{LR}(p)}
\exp\!\left[
\frac{
\left\langle Q_{HR}^{p}, K_{LR}^{q'}\right\rangle
}{
\sqrt{d_{enc}}
}
\right]
\end{equation}

\subsection{Semantic Query-Key Branch}
\label{sec:sqk}
The semantic query key branch utilizes image $\mathbf{I}$ to obtain semantic encoding $E_{sem}$ from a fusion of sources (i) a frozen Semantic Encoder (DINOv3 ConvNeXt-S) and (ii) a lightweight Pixel Encoder via Cross-Attention Chain. This setup allows the resulting query-key pair to be semantically aware, high in resolution, and independent of the VFM.


\paragraph{Pixel Encoder.} The Pixel Encoder is a shallow convolutional module comprising a $3{\times}3$ convolution followed by four $1{\times}1$ residual blocks at width $d_{\mathrm{enc}}$. From the raw RGB image, the encoder extracts local texture features and produces the initial Cross-Attention Chain query $\theta_0$ at $H/8 \times W/8$. At this stage, $\theta_0$ captures local context but lacks global semantics.

\paragraph{Semantic Encoder.} The global semantic guidance $E_{\mathrm{sem}}$ is extracted by a frozen DINOv3 ConvNeXt-S encoder, whose hierarchical architecture produces four semantically rich feature maps $S_3,\ldots,S_0$ at resolutions from (H/2) to (H/16). The encoder remains frozen during self-supervised training and introduces no trainable parameters.



\paragraph{Cross-Attention Chain.} Inspired by the top-down pathway of feature pyramid networks~\cite{fpn}, the chain fuses the multi-scale pyramid semantic features $S_{N}$ into the query $\theta_{N}$ gradually from coarsest to finest. This fusion gradually increase in spatial resolution and simultaneously accumulates semantics at each stage. A  single cross-attention chain consists of three steps. First, $S_{N}$ from semantic encoder is upsampled by nearest-neighbor interpolation followed by a $3{\times}3$ convolution, denoted $S^{N}_{up}$, and added to the query, giving the intermediate $\tilde{\theta}_{N}=\theta_{N}+S^{N}_{up}$. Second, $\tilde{\theta}_{N}$ is normalized and projected as the attention query, while $S_{N}$ is normalized and projected as the keys and values. The key and value projection maps $S_{N}$ from its native feature width onto the shared feature width $d_{enc}$ while aggregating semantic information from every scale. Third, neighborhood attention followed by a feed-forward layer produces $Out^{N}_{FF}$, which is added back as residual. The query is then updated according to Equation \ref{eq:chain}.

\begin{equation} 
\theta_{(N+1)} \;\leftarrow\; \theta_{N} \;+\; S^{N}_{up} \;+\; Out^{N}_{FF}. 
\label{eq:chain} 
\end{equation}

This chain is initially applied to $\theta_{0}$, produced by the Pixel Encoder, together with $S_{0}$, to obtain $\theta_{1}$. Similarly, each subsequent stage of cross-attention chain doubles the spatial resolution which enhances the query scale from $\theta_{1}$ at $H/8\times W/8$ to $\theta_{4}$ at the full $H\times W$ grid. The final stage draws on $S_{3}$, upsampled by \emph{NN$+$Conv2D} block, is added directly without an attention term, for two reasons. (i) Attending at $H\times W$ would be the most expensive read in the chain and (ii) $S_{3}$ originates from the earliest layers of the Semantic Encoder and carries the least semantic content. This direct addition provides low-level structural detail rather than semantic meaning. This final query $\theta_{4}$ is the semantic encoding $E_{sem}$ which carries both appearance and semantics and eventually produces the query $Q_{HR}$ and the key $K_{LR}$ in \ours{} architecture (see Figure \ref{fig:architecture}).

\paragraph{Self-supervised training.} \ours{} is trained in self-supervised manner, similar to recent upsamplers \cite{naf,jafar}. Given a high-resolution image $I$, we form a low-resolution view $I_{LR}$ by bilinearly downsampling it by a random factor ($0.3, 0.5, 0.7$). Passing $I$ through the frozen teacher yields the target $F_{HR}$, while passing $I_{LR}$ through the same teacher yields the low-resolution input $F_{LR}$. \ours{} upsamples the latter under the guidance of the full-resolution image, $\hat{F}_{HR} := \ours{}(I, F_{LR})$, and is supervised against $F_{HR}$ by a mean-squared reconstruction loss $\mathcal{L}_{MSE} = \frac{1}{N_{HR}}\sum_{n} \big\lVert \hat{F}_{HR}^{(n)} - F_{HR}^{(n)} \big\rVert_2^2$, where the sum runs over all $N_{HR}$ high-resolution feature locations. We optimise with AdamW (learning rate $2{\times}10^{-4}$, weight decay $0.05$), a cosine schedule with $1{,}000$ warm-up steps, gradient clipping at $1.0$, batch size $2$, and bf16 precision, for $50{,}000$ steps on a mixture of SA-1B \cite{sa1b}, and COCO \cite{coco} sampled uniformly. The model uses encoder width $d_{enc}{=}256$ ($6.27$M trainable parameters).

\begin{figure}
\centering
\includegraphics[width=1\linewidth]{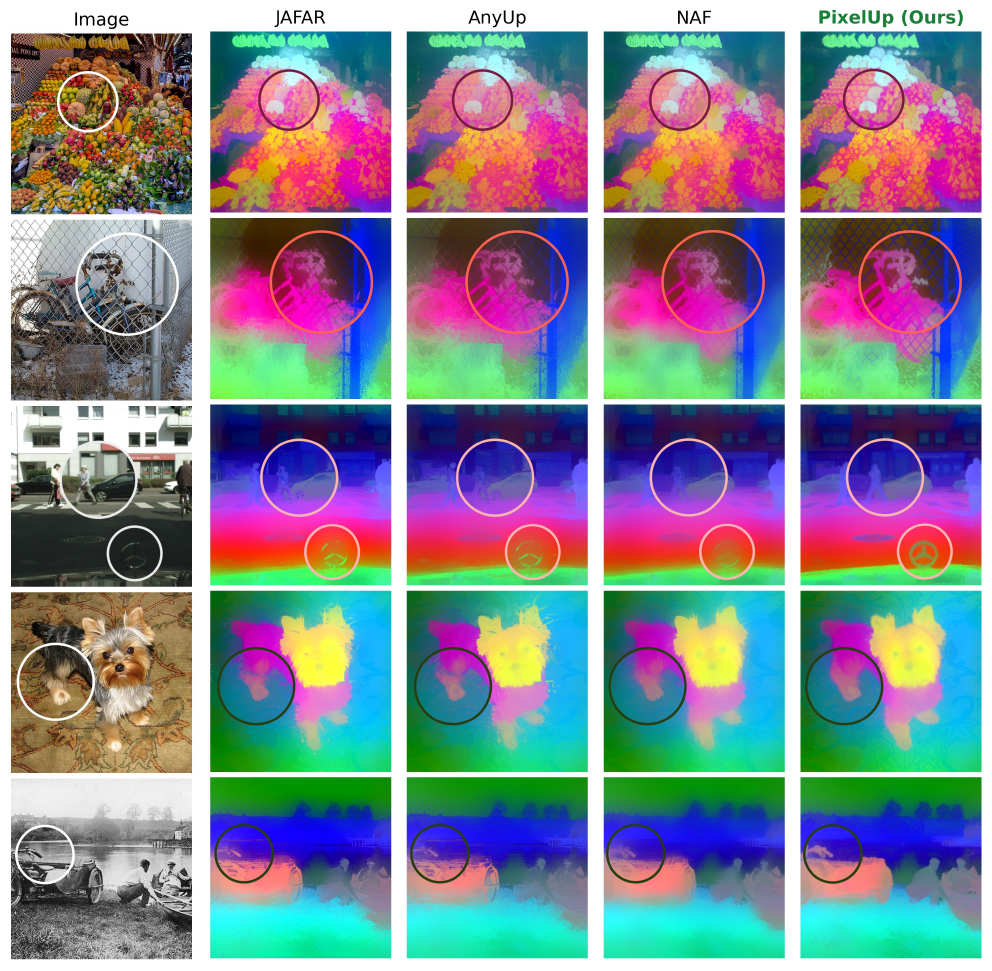}
\caption{\textbf{Qualitative comparison across feature upsamplers.} The encircled region highlights the artifacts that other upsamplers exhibit, while ours yields the most coherent semantic maps with clean object boundaries.}
\label{fig:qualitative}
\end{figure}

\section{Experiments}

\begin{table*}[t]
\centering
{\small
\setlength{\tabcolsep}{0.5pt}
\renewcommand{\arraystretch}{1.2}
\begin{tabular}{@{}l cccc cc c c cccc cc c c cccc cc c c@{}}
\toprule
\multirow{3}{*}{\textbf{Method}} & \multicolumn{8}{c}{\textbf{\textit{Cityscapes}} (mIoU$\uparrow$)} & \multicolumn{8}{c}{\textbf{\textit{ADE20K}} (mIoU$\uparrow$)} & \multicolumn{8}{c}{\textbf{\textit{VOC}} (mIoU$\uparrow$)} \\
\cmidrule(lr){2-9}\cmidrule(lr){10-17}\cmidrule(lr){18-25}
 & \multicolumn{4}{c}{\textbf{DINO}} & \multicolumn{2}{c}{\textbf{RADIO}} & \multirow{2}{*}{\textbf{Fr.}} & \multirow{2}{*}{\textbf{$\Delta_{nn}$}} & \multicolumn{4}{c}{\textbf{DINO}} & \multicolumn{2}{c}{\textbf{RADIO}} & \multirow{2}{*}{\textbf{Fr.}} & \multirow{2}{*}{\textbf{$\Delta_{nn}$}} & \multicolumn{4}{c}{\textbf{DINO}} & \multicolumn{2}{c}{\textbf{RADIO}} & \multirow{2}{*}{\textbf{Fr.}} & \multirow{2}{*}{\textbf{$\Delta_{nn}$}} \\
\cmidrule(lr){2-5}\cmidrule(lr){6-7}\cmidrule(lr){10-13}\cmidrule(lr){14-15}\cmidrule(lr){18-21}\cmidrule(lr){22-23}
 & \textbf{v3-S} & \textbf{v3-B} & \textbf{v3-L} & \textbf{v2-R} & \textbf{v2.5} & \textbf{v4} & & & \textbf{v3-S} & \textbf{v3-B} & \textbf{v3-L} & \textbf{v2-R} & \textbf{v2.5} & \textbf{v4} & & & \textbf{v3-S} & \textbf{v3-B} & \textbf{v3-L} & \textbf{v2-R} & \textbf{v2.5} & \textbf{v4} & & \\
\midrule
Nearest & 51.4 & 58.8 & 59.9 & 56.5 & 53.2 & 59.8 & 54.5 & $-$ & 37.0 & 43.6 & 45.3 & 40.7 & 41.4 & 47.5 & 36.6 & $-$ & 80.6 & 85.0 & 85.3 & 79.7 & 81.7 & 86.2 & 79.5 & $-$ \\
Bilinear & 53.9 & 63.0 & 64.8 & 61.7 & 57.7 & 65.5 & 59.0 & \textbf{\textcolor{green!45!black}{$+$4.5}} & 38.1 & 45.5 & 47.4 & 43.5 & 43.9 & 50.1 & 38.9 & \textbf{\textcolor{green!45!black}{$+$2.2}} & 82.2 & 86.9 & 87.5 & 83.1 & 84.4 & 88.8 & 82.3 & \textbf{\textcolor{green!45!black}{$+$2.5}} \\
JBU & 54.6 & 63.8 & 65.6 & 63.1 & 58.8 & 66.4 & 60.4 & \textbf{\textcolor{green!45!black}{$+$5.5}} & 38.3 & 45.8 & 47.7 & 44.2 & 44.3 & 50.5 & 39.3 & \textbf{\textcolor{green!45!black}{$+$2.6}} & 82.3 & 87.1 & 87.9 & 83.7 & 84.9 & 89.1 & 82.7 & \textbf{\textcolor{green!45!black}{$+$2.8}} \\
JAFAR & 53.4 & 62.1 & 63.5 & -- & 57.4 & -- & -- & \textbf{\textcolor{green!45!black}{$+$3.2}} & 38.3 & 45.8 & 47.6 & -- & 44.9 & -- & -- & \textbf{\textcolor{green!45!black}{$+$2.3}} & 82.2 & 87.0 & 87.6 & -- & 85.6 & -- & -- & \textbf{\textcolor{green!45!black}{$+$2.5}} \\
AnyUp & 51.3 & 60.3 & 62.8 & 61.7 & 56.9 & 62.8 & 58.4 & \textbf{\textcolor{green!45!black}{$+$2.9}} & 37.9 & 45.2 & 47.2 & 45.3 & 44.6 & 49.8 & 39.9 & \textbf{\textcolor{green!45!black}{$+$2.5}} & 82.0 & 86.6 & 87.5 & 85.5 & 85.6 & 88.5 & 83.2 & \textbf{\textcolor{green!45!black}{$+$3.0}} \\
NAF & \underline{55.6} & \underline{65.0} & \underline{66.7} & \underline{66.3} & \underline{61.0} & \underline{67.6} & \underline{62.7} & \textbf{\textcolor{green!45!black}{$+$7.2}} & \underline{39.0} & \underline{46.5} & \underline{48.3} & \textbf{46.4} & \underline{45.8} & \underline{51.1} & \underline{40.6} & \textbf{\textcolor{green!45!black}{$+$3.6}} & \underline{83.0} & \underline{87.8} & \underline{88.6} & \underline{86.3} & \underline{86.7} & \underline{89.9} & \underline{84.3} & \textbf{\textcolor{green!45!black}{$+$4.1}} \\
\rowcolor{blue!8} \textbf{\ours{}} & \textbf{58.3} & \textbf{68.2} & \textbf{70.1} & \textbf{68.3} & \textbf{63.4} & \textbf{70.2} & \textbf{64.3} & \textbf{\textcolor{green!45!black}{$+$9.8}} & \textbf{39.6} & \textbf{47.3} & \textbf{49.1} & \underline{46.3} & \textbf{46.1} & \textbf{51.9} & \textbf{40.7} & \textbf{\textcolor{green!45!black}{$+$4.1}} & \textbf{84.0} & \textbf{88.7} & \textbf{89.4} & \textbf{86.4} & \textbf{87.0} & \textbf{90.5} & \textbf{84.3} & \textbf{\textcolor{green!45!black}{$+$4.6}} \\
\midrule
\textbf{\textit{$\Delta_{\text{NAF}}$}} & \textbf{\textcolor{green!45!black}{$+$2.7}} & \textbf{\textcolor{green!45!black}{$+$3.2}} & \textbf{\textcolor{green!45!black}{$+$3.4}} & \textbf{\textcolor{green!45!black}{$+$2.0}} & \textbf{\textcolor{green!45!black}{$+$2.4}} & \textbf{\textcolor{green!45!black}{$+$2.6}} & \textbf{\textcolor{green!45!black}{$+$1.7}} & $-$ & \textbf{\textcolor{green!45!black}{$+$0.6}} & \textbf{\textcolor{green!45!black}{$+$0.8}} & \textbf{\textcolor{green!45!black}{$+$0.8}} & \textbf{\textcolor{red!65!black}{$-$0.1}} & \textbf{\textcolor{green!45!black}{$+$0.4}} & \textbf{\textcolor{green!45!black}{$+$0.8}} & \textbf{\textcolor{green!45!black}{$+$0.1}} & $-$ & \textbf{\textcolor{green!45!black}{$+$1.0}} & \textbf{\textcolor{green!45!black}{$+$0.9}} & \textbf{\textcolor{green!45!black}{$+$0.8}} & \textbf{\textcolor{green!45!black}{$+$0.1}} & \textbf{\textcolor{green!45!black}{$+$0.3}} & \textbf{\textcolor{green!45!black}{$+$0.7}} & \textbf{\textcolor{green!45!black}{$+$0.0}} & $-$ \\
\bottomrule
\end{tabular}
}
\caption{\textbf{Semantic segmentation} (mIoU $\uparrow$, linear probe), three datasets side by side; backbones grouped by family. \ours{} (shaded) is one frozen checkpoint. \textbf{Bold} best, \underline{underline} second per column; $\Delta_{nn}$ is mean gain over Nearest (JAFAR gain is over its 4 covered backbones only); $\Delta_{\text{NAF}}$ (bottom row) is \ours{} minus NAF per backbone (\textcolor{red!65!black}{red}$=$loss). Fr.=Franca.}
\label{tab:seg-transpose}
\end{table*}

\subsection{Experimental Setup}
We evaluate \ours{} under qualitative, supervised, and training-free protocols that test spatial coherence, downstream feature quality, and compatibility with fixed pipelines. For qualitative analysis, each upsampled feature map is projected to RGB using a PCA basis shared across methods within each image row. We assess boundary separation and color consistency within visible object regions rather than assigning semantic meaning to individual PCA colors. For supervised evaluation, the backbone and upsampler remain frozen while a lightweight $1{\times}1$ prediction head is trained for segmentation or monocular depth estimation following same training settings as \cite{naf}. For training-free evaluation, upsampled features replace bilinear interpolation within fixed DiffCut~\cite{diffcut} and RADSeg~\cite{radseg} methods for unsupervised and open-vocabulary segmentation tasks. Together, the protocols test whether one frozen \ours{} checkpoint transfers across VFMs and produces high-fidelity features for supervised segmentation and depth prediction and training-free segmentation tasks. 

\paragraph{Backbones, datasets, and metrics.}
We evaluate one frozen \ours{} checkpoint across ten target VFMs, five datasets, and four task settings. We compare against existing methods, including Nearest, Bilinear, Joint Bilateral Upsampling (JBU)~\cite{jbu}, JAFAR~\cite{jafar}, AnyUp~\cite{anyup}, and NAF~\cite{naf}. Supervised probing uses DINOv3-S/B/L/7B~\cite{dinov3}, DINOv2-R-B~\cite{dinov2}, RADIOv2.5-B and RADIOv4~\cite{radio}, and Franca-B~\cite{franca}. Their feature widths range from $384$ to $4096$, with patch strides of $14$ or $16$. Training-free evaluation uses SSD-1B~\cite{ssd1b} for DiffCut and C-RADIOv3-B~\cite{radio}, paired with SigLIP2~\cite{siglip2}, for RADSeg. We evaluate supervised semantic segmentation on Pascal VOC~\cite{voc}, Cityscapes~\cite{cityscapes}, and ADE20K~\cite{ade20k}, and monocular depth estimation on NYUv2~\cite{nyuv2}. Training-free segmentation uses ADE20K, Cityscapes, KITTI~\cite{kitti}, and Pascal VOC. We report mean Intersection-over-Union for all segmentation tasks. For depth, we report $\delta_1$, the percentage of valid pixels whose predicted-to-ground-truth depth ratio lies within $1.25$, following Probe3D~\cite{deptheval}. The evaluation examines transfer across backbone families, feature dimensions, datasets, and DINOv3 model scales from Small to 7B. We use respective official released repositories or checkpoints for all compared methods.

\section{Results and Discussion}

\subsection{Qualitative Analysis}
Figure~\ref{fig:qualitative} projects the upsampled features of DINOv3-L onto a shared three-component PCA basis per row. Across all examples, \ours{} yields the most object-coherent maps, with uniform interior color and clean object boundaries without drift toward image textures, while other baselines leak texture around boundaries, particularly where foreground and background share similar texture (see encircled regions).  The same behavior drives the quantitative results presented below.

\subsection{Supervised Linear Probing}
\subsubsection{Semantic Segmentation.}

Table ~\ref{tab:seg-transpose} summarizes the semantic segmentation results. Across all three datasets, \ours{} outperforms both VFM-agnostic baselines and the VFM-specific JAFAR. Compared with JAFAR, \ours{} improves mIoU by an average of $+5.9$, $+1.4$, and $+1.7$ on Cityscapes, ADE20K, and VOC, respectively, winning all four shared backbones on every dataset. Compared with the VFM-agnostic SOTA NAF, \ours{} gains an average of $+2.5$, $+0.5$, and $+0.5$ mIoU across seven backbones, winning 20 of 21 comparisons; the only loss is by $0.1$ mIoU on DINOv2-R (ADE20K).

We evaluate \ours{} \textbf{across VFM families} including distilled (RADIOv2.5/v4), self-supervised (DINOv3-S/B/L, DINOv2-R), and Franca model. \ours{} consistently improves over the VFM-agnostic SOTA NAF across every family, with mean gains of $+1.58$ mIoU on DINOv3, $+1.17$ on RADIO, $+0.67$ on DINOv2-R, and $+0.57$ on Franca-B. \textbf{Across model sizes}, \ours{} performs consistently across the DINOv3 hierarchy. On Cityscapes, \ours{} surpasses JAFAR by +4.9, +6.1, and +6.6 mIoU with DINOv3-S, DINOv3-B, and DINOv3-L, respectively, and exceeds NAF by +2.7, +3.2, and +3.4 mIoU. This shows that the learned upsampling rule scales robustly across model sizes.

The granularity of the target classes varies \textbf{across datasets}. Measured on the validation splits, VOC averages $1.7$  foreground classes per image with a median foreground area of $23\%$, ADE20K $8.3$ classes with a median area of $4.8\%$, and Cityscapes $12.3$ classes, where thin structures like traffic lights cover $0.19\%$ of pixels and poles $1.6\%$. On the coarse VOC benchmark, where the patch tokens already cover the large foreground objects, every upsampler improves over bilinear by only $+0.3$ to $+2.2$ mIoU. On Cityscapes, however, JAFAR and AnyUp fall below bilinear by $-0.75$ and $-1.63$ mIoU on average. Only JBU ($+1.0$), NAF ($+2.8$), and \ours{} ($+5.3$) improve over bilinear.

\subsubsection{Monocular Depth.}
Table ~\ref{tab:depth} reports monocular depth estimation on NYUv2. \ours{} achieves the best $\delta_1$ on five of six backbones and trailing NAF by only $0.01$ on RADIOv2.5 with mean improvement of $+0.25$ $\delta_1$ over NAF across VFMs.  \textbf{Across model sizes}, \ours{} transfers consistently from DINOv3-S to the 7B model. It achieves the highest $\delta_1$ on all four DINOv3 scales, with gains over NAF of $+0.17$, $+0.18$, $+0.14$, and $+0.77$ on DINOv3-S/B/L/7B, respectively (Table \ref{tab:depth}). Notably, unlike JAFAR, which runs out of GPU memory on DINOv3-7B even with batch size 1, \ours{} scales to the 7B backbone without training changes. 

\begin{table}
\centering
{\small
\setlength{\tabcolsep}{2pt}
\renewcommand{\arraystretch}{1.1}
\begin{tabular}{@{}l cccc cc c@{}}
\toprule
\multirow{2}{*}{\textbf{Method}} & \multicolumn{4}{c}{\textbf{DINO}} & \multicolumn{2}{c}{\textbf{RADIO}} & \multirow{2}{*}{$\Delta_{nn}$} \\
\cmidrule(lr){2-5}\cmidrule(lr){6-7}
 & \textbf{v3-S} & \textbf{v3-B} & \textbf{v3-L} & \textbf{v3-7B} & \textbf{v2.5} & \textbf{v4} & \\
\midrule
Nearest         & 82.83 & 87.06 & 87.08 & 89.92 & 84.31 & 92.49 & -- \\
Bilinear        & 83.52 & 87.90 & 88.51 & 90.69 & 85.76 & 93.44 & \textbf{\textcolor{green!45!black}{$+$1.07}} \\
JAFAR           & 83.42 & 88.21 & \underline{89.18} & \textcolor{red!65!black}{OOM} & 86.21 & --    & \textbf{\textcolor{green!45!black}{$+$1.43}} \\
AnyUp           & 83.53 & 88.16 & 89.06 & \underline{93.43} & 86.48 & 93.41 & \textbf{\textcolor{green!45!black}{$+$1.37}} \\
NAF             & \underline{83.76} & \underline{88.32} & 89.11 & 93.08 & \textbf{86.71} & \underline{93.72} & \textbf{\textcolor{green!45!black}{$+$1.57}} \\
\rowcolor{blue!8}
\textbf{\ours{}} & \textbf{83.93} & \textbf{88.50} & \textbf{89.25} & \textbf{93.85} & \underline{86.70} & \textbf{93.97} & \textbf{\textcolor{green!45!black}{$+$1.72}} \\
\midrule
\textbf{\textit{$\Delta_{\text{NAF}}$}}
  & \textbf{\textcolor{green!45!black}{$+$0.17}}
  & \textbf{\textcolor{green!45!black}{$+$0.18}}
  & \textbf{\textcolor{green!45!black}{$+$0.14}}
  & \textbf{\textcolor{green!45!black}{$+$0.77}}
  & \textbf{\textcolor{red!65!black}{$-$0.01}}
  & \textbf{\textcolor{green!45!black}{$+$0.25}}
  & \textbf{\textcolor{green!45!black}{$+$0.25}} \\
\bottomrule
\end{tabular}
}
\caption{\textbf{Monocular depth estimation on NYUv2} ($\delta_{1}\uparrow$): Results across backbones including the DINOv$3$-$7$B model. \textbf{Bold} best, \underline{underline} second per column. $\Delta_{nn}$ is calculated against Nearest. $\Delta_{\text{NAF}}$ is the improvement over NAF.}
\label{tab:depth}
\end{table}

\subsection{Training-free Evaluation}

\paragraph{Unsupervised Segmentation.}

\begin{table}
\centering
{\small
\setlength{\tabcolsep}{3.5pt}
\renewcommand{\arraystretch}{1.1}
\begin{tabular}{@{}l cccc c c@{}}
\toprule
\textbf{Method} & \textbf{ADE} & \textbf{Cityscapes} & \textbf{KITTI} & \textbf{VOC} & \textbf{\textit{Avg}} & $\Delta_{nn}$ \\
\midrule
\multicolumn{7}{@{}c}{\textbf{\textit{(a) Unsupervised (DiffCut, SSD-1B)}}} \\
\midrule
Nearest & 44.44 & 30.47 & \underline{33.46} & 65.15 & 43.38 & $-$ \\
Bilinear & 44.37 & 30.50 & 33.44 & 65.25 & 43.39 & \textbf{\textcolor{green!45!black}{$+$0.01}} \\
AnyUp & 44.38 & 30.95 & 33.09 & \underline{65.30} & 43.43 & \textbf{\textcolor{green!45!black}{$+$0.05}} \\
NAF & \underline{44.49} & \underline{31.00} & 33.40 & 65.21 & \underline{43.53} & \textbf{\textcolor{green!45!black}{$+$0.15}} \\
\rowcolor{blue!8}
\textbf{\ours{}} & \textbf{44.55} & \textbf{31.45} & \textbf{33.97} & \textbf{65.96} & \textbf{43.98} & \textbf{\textcolor{green!45!black}{$+$0.60}} \\
\textbf{\textit{$\Delta_{\text{NAF}}$}} & \textbf{\textcolor{green!45!black}{$+$0.06}} & \textbf{\textcolor{green!45!black}{$+$0.45}} & \textbf{\textcolor{green!45!black}{$+$0.57}} & \textbf{\textcolor{green!45!black}{$+$0.75}} & \textbf{\textcolor{green!45!black}{$+$0.45}} & $-$ \\
\midrule
\multicolumn{7}{@{}c}{\textbf{\textit{(b) Open-vocabulary (RADSeg, RADIOv3-B)}}} \\
\midrule
Nearest & 27.30 & 40.77 & 40.29 & 85.98 & 48.59 & $-$ \\
Bilinear & 28.52 & 42.32 & 41.93 & 88.21 & 50.25 & \textbf{\textcolor{green!45!black}{$+$1.66}} \\
AnyUp & \textcolor{red!65!black}{OOM} & \textcolor{red!65!black}{OOM} & \textcolor{red!65!black}{OOM} & \textbf{89.50} & -- & $-$ \\
NAF & \textbf{30.56} & \underline{43.42} & \underline{44.37} & \underline{89.49} & \underline{51.96} & \textbf{\textcolor{green!45!black}{$+$3.37}} \\
\rowcolor{blue!8}
\textbf{\ours{}} & \underline{30.17} & \textbf{46.71} & \textbf{47.04} & 89.25 & \textbf{53.29} & \textbf{\textcolor{green!45!black}{$+$4.70}} \\
\textbf{\textit{$\Delta_{\text{NAF}}$}} & \textbf{\textcolor{red!65!black}{$-$0.39}} & \textbf{\textcolor{green!45!black}{$+$3.29}} & \textbf{\textcolor{green!45!black}{$+$2.67}} & \textbf{\textcolor{red!65!black}{$-$0.24}} & \textbf{\textcolor{green!45!black}{$+$1.33}} & $-$ \\
\bottomrule
\end{tabular}}
\caption{\textbf{Training-free segmentation} (mIoU$\uparrow$): (a) Unsupervised with DiffCut~\cite{diffcut} on SSD-1B and (b) open-vocabulary with RADSeg on RADIOv3-B \textbf{Bold} best, \underline{underline} second per column. $\Delta_{nn}$ is calculated against Nearest. $\Delta_{\text{NAF}}$ is the improvement over NAF.}
\label{tab:trainingfree}
\end{table}
On DiffCut~\cite{diffcut} with SSD-1B features (Table \ref{tab:trainingfree}), \ours{} leads on every benchmark at the $128$ mask grid, reaching $43.98$~mIoU on average, $+0.46$ over NAF and $+0.59$ over Bilinear. DiffCut partitions the feature affinity graph by recursive normalized cuts, so mask quality tracks upsampled feature fidelity. The margin widens with the output grid. AnyUp exhausts memory at the $512$ grid and NAF at $1024$, while \ours{} still operates at both. The supplementary material reports the full resolution sweep.

\paragraph{Open-Vocabulary Segmentation.}
In the RADSeg~\cite{radseg} pipeline (Table \ref{tab:trainingfree}), \ours{} achieves the best average mIoU at $53.29$, $+1.33$ over NAF, with the largest gains on the thin-structure benchmarks Cityscapes ($+3.29$) and KITTI ($+2.67$). \ours{} loses by a small margin on ADE20K ($-0.39$ vs.\ NAF) and ranks third on the near-saturated VOC, behind AnyUp and NAF by $0.25$~mIoU. AnyUp exhausts memory on three of the four benchmarks, while \ours{} runs on all four benchmark datasets.
\section{Efficiency and Memory}

Table \ref{tab:efficiency} reports throughput (frames per second) and peak allocated memory for $448^2$ and $1024^2$ outputs on backbones of width $768$ (DINOv3-B) and $4096$ (DINOv3-7B), measured on a single $48$~GB L40S GPU at bf16 precision. \ours{} is the fastest learned upsampler and uses the lowest peak memory across both settings. It is also the only learned upsampler that fits DINOv3-7B at $1024^2$ output. Across both backbones, \ours{} runs $1.6\times$--$1.7\times$ faster than NAF and uses $1.6\times$--$1.9\times$ less memory, while JAFAR and AnyUp exhaust memory on DINOv3-7B.

\begin{table}[t]
\centering
{\small
\setlength{\tabcolsep}{1.9pt}
\renewcommand{\arraystretch}{1.15}
\begin{tabular}{@{}l cc cc cc cc @{}}
\toprule
& \multicolumn{4}{c}{\textbf{\textit{Output $448^2$}}} & \multicolumn{4}{c}{\textbf{\textit{Output $1024^2$}}} \\
\cmidrule(lr){2-5}\cmidrule(lr){6-9}
& \multicolumn{2}{c}{\textbf{768-d}} & \multicolumn{2}{c}{\textbf{4096-d}} & \multicolumn{2}{c}{\textbf{768-d}} & \multicolumn{2}{c}{\textbf{4096-d}} \\
\cmidrule(lr){2-3}\cmidrule(lr){4-5}\cmidrule(lr){6-7}\cmidrule(lr){8-9}
\textbf{Method} & \textbf{FPS} & \textbf{GB} & \textbf{FPS} & \textbf{GB} & \textbf{FPS} & \textbf{GB} & \textbf{FPS} & \textbf{GB} \\
\midrule
JAFAR   & 12.3 & 7.67 & 11.6 & 7.68 & \multicolumn{2}{c}{\textcolor{red!65!black}{OOM}} & \multicolumn{2}{c}{\textcolor{red!65!black}{OOM}} \\
AnyUp   &  9.5 & 7.57 &  7.6 & 7.58 & \multicolumn{2}{c}{\textcolor{red!65!black}{OOM}} & \multicolumn{2}{c}{\textcolor{red!65!black}{OOM}} \\
NAF     & 19.5 & 2.79 &  9.3 & 10.81 & 2.3 & 14.60 & \multicolumn{2}{c}{\textcolor{red!65!black}{OOM}} \\
\rowcolor{blue!8} \ours{} & \textbf{32.3} & \textbf{1.77} & \textbf{15.2} & \textbf{5.64} & \textbf{4.0} & \textbf{8.67} & \textbf{2.1} & \textbf{28.88} \\
\textbf{\textit{$\Delta_{\text{NAF}}$}} & \textbf{\textcolor{green!45!black}{1.66\texttimes}} & \textbf{\textcolor{green!45!black}{1.58\texttimes}} & \textbf{\textcolor{green!45!black}{1.63\texttimes}} & \textbf{\textcolor{green!45!black}{1.92\texttimes}} & \textbf{\textcolor{green!45!black}{1.74\texttimes}} & \textbf{\textcolor{green!45!black}{1.69\texttimes}} & $-$ & $-$ \\
\bottomrule
\end{tabular}
}
\caption{\textbf{Efficiency and peak memory.} FPS is frames per second; GB is peak allocated memory.}
\label{tab:efficiency}
\end{table}
\section{Ablations}
\label{sec:abl}
We ablate one design choice at a time. Results are reported in Table \ref{tab:ablation}, where means average VOC, Cityscapes, and KITTI is measured with fixed base anchor ($d_{\text{enc}}{=}384$, heads $6/6$, windowed, cross-attention, nn-conv). Across ablations the deployed configuration is shaded. Three observations follow: (i) Setting aside encoder width, guidance, and CARAFE, every variant sits within $0.077$ mean mIoU of the base, below noise. (ii) Accuracy plateaus with increase in width, hence the deployed encoder at $d_{\text{enc}}{=}256$ is chosen for a $43\%$ parameter reduction without accuracy loss. (iii) Removing the Semantic guidance drops accuracy by $3.32$ mean mIoU \emph{and} slows throughput, because without a coarse pyramid to lift, the Pixel Encoder must produce the query at full resolution. Where accuracy is flat we select on cost, keeping windowed attention over global for throughput, nn-conv over pixel-shuffle for parameter count, and cross-attention over concatenation.
\begin{table}[t]
\centering
{\small
\setlength{\tabcolsep}{2pt}
\renewcommand{\arraystretch}{1}
\begin{tabular}{@{}l cc cc cc@{}}
\toprule
& \multicolumn{2}{c}{\textit{Linear probing}} & \textit{Open-vocab} & & & \\
\cmidrule(lr){2-3}\cmidrule(lr){4-4}
Variant & VOC & CS & KITTI & \textit{Mean} & Par. & FPS \\
\midrule
\multicolumn{7}{@{}l}{\textit{Encoder width $d_{\text{enc}}$}}\\
128 & 88.72 & 68.34 & 45.91 & 67.66 & 2.59 & 44.6 \\
192 & 88.78 & 68.98 & 46.61 & 68.12 & 4.29 & 36.8 \\
\rowcolor{blue!8} 256 & 88.75 & 69.12 & 47.04 & 68.30 & 6.27 & 32.3 \\
384 & 88.73 & 69.28 & 46.92 & \textbf{68.31} & 11.07 & 23.9 \\
\midrule
\multicolumn{7}{@{}l}{\textit{Neighborhood attention}}\\
global & 88.66 & 69.26 & 47.04 & \textbf{68.32} & 11.07 & 22.7 \\
\rowcolor{blue!8} windowed & 88.73 & 69.28 & 46.92 & 68.31 & 11.07 & 23.9 \\
\midrule
\multicolumn{7}{@{}l}{\textit{Encoder heads}}\\
4 & 88.76 & 69.27 & 46.91 & \textbf{68.31} & 11.07 & 23.6 \\
\rowcolor{blue!8} 6 & 88.73 & 69.28 & 46.92 & \textbf{68.31} & 11.07 & 23.9 \\
12 & 88.72 & 69.27 & 46.88 & 68.29 & 11.07 & 24.2 \\
\midrule
\multicolumn{7}{@{}l}{\textit{Decoder heads}}\\
4 & 88.75 & 69.27 & 46.99 & 68.34 & 11.07 & 25.1 \\
\rowcolor{blue!8} 6 & 88.73 & 69.28 & 46.92 & 68.31 & 11.07 & 23.9 \\
12 & 88.70 & 69.27 & 47.19 & \textbf{68.39} & 11.07 & 21.8 \\
\midrule
\multicolumn{7}{@{}l}{\textit{Guidance fusion}}\\
concatenation & 88.64 & 69.15 & 47.05 & 68.28 & 7.24 & 26.2 \\
\rowcolor{blue!8} cross-attention & 88.73 & 69.28 & 46.92 & \textbf{68.31} & 11.07 & 23.9 \\
no fusion & 87.25 & 63.84 & 43.87 & 64.99 & 1.08 & 4.8 \\
\midrule
\multicolumn{7}{@{}l}{\textit{Upsampling operator}}\\
pixel-shuffle & 88.61 & 69.03 & 47.16 & 68.27 & 26.00 & 24.2 \\
CARAFE & 88.52 & 68.73 & 47.15 & 68.13 & 6.97 & 21.6 \\
\rowcolor{blue!8} nn-conv & 88.73 & 69.28 & 46.92 & \textbf{68.31} & 11.07 & 23.9 \\
\bottomrule
\end{tabular}
}
\caption[Architecture ablations]{\textbf{Architecture ablations.} Deployed configuration shaded, \textbf{bold} marks the block-best mean. Par.\ is trainable parameters (M); FPS at DINOv3-B, $448^2$.}
\label{tab:ablation}
\end{table}


\section{Conclusion}

We introduced \ours{}, a zero-shot, semantic feature upsampling method that upsamples patch-token features of any VFM to recover high-fidelity semantic feature at pixel resolution.  Existing attention-based upsamplers lack semantic guidance for the query, resulting in textural artifacts at pixel resolution and degraded performance on fine-grained tasks. \ours{} addresses this limitation by drawing semantic guidance into the query from multi-scale semantic features of a pre-trained frozen Semantic Encoder. We evaluate \ours{} across ten VFMs, including models up to 7B parameters and show that zero-shot transfers performs across architectures, feature dimensions, and model scales. Across supervised tasks, \ours{} leads in 20 of 21 segmentation settings and five of six NYUv2 depth settings. Improvements of $+0.5$ and $+1.3$ mIoU for unsupervised and open-vocabulary segmentation task further demonstrate compatibility with training-free pipelines. Under matched settings, \ours{} also provides $1.6{\times}$--$1.7{\times}$ higher throughput and $1.6{\times}$--$1.9{\times}$ lower peak memory than prior SOTA. These findings show that pretrained semantic guidance can improve dense feature reconstruction without sacrificing cross-backbone transfer or computational efficiency. Future work will investigate if the same principle applies to specialized visual domains and remains effective with alternative frozen semantic encoders.

This work was performed at the University of Houston under a contract with the Commercial Smallsat Data Scientific Analysis Program of NASA (NNH22ZDA001N-CSDSA) and the NASA Decadal Survey Incubation Program: Science and Technology (NNH21ZDA001N-DSI). 

\bibliography{pixelup}


\end{document}


\maketitle

\setcounter{tocdepth}{3}
\tableofcontents
\vspace{1em}

\section{Related Work and Preliminaries}
This section expands the main paper's related work, and Sec.~\ref{sec:attn-upsamplers} discusses the attention-based upsamplers in detail.

\subsection{Classical Filters and Task-Specific Operators}
Early feature upsamplers fall into two groups, and neither reconstructs the semantics of a vision
foundation model (VFM) the way the attention-based methods of Sec.~\ref{sec:attn-upsamplers} do.
Training-free filters read only photometric cues, so on VFM features they track texture
and color rather than content: Joint Bilateral Upsampling weights each output pixel by color
similarity to the full-resolution guide image~\cite{jbu}, and the guided filter replaces its
Gaussian range weights with a closed-form local linear model~\cite{guidedfilter}. Task-specific
operators (CARAFE~\cite{carafe}, FADE~\cite{fade}, SAPA~\cite{sapa}, DySample~\cite{dysample})
instead learn reassembly kernels or sampling offsets end-to-end inside one encoder-decoder
network, under that task's supervision and at a small fixed factor per stage.
\subsection{Attention-Based Upsamplers}
\label{sec:attn-upsamplers}
JAFAR~\cite{jafar}, AnyUp~\cite{anyup} and NAF~\cite{naf} are the closest prior work. All three
cast upsampling as cross-attention in which the target VFM supplies the values, and all three
serve an output grid of any size from one trained model. Two design choices make this work. The
query lives at the requested output resolution, so the output grid is an argument rather than a
fixed multiple of the input. The key is pooled to the low-resolution feature grid, so query-key
attention scores relate output positions to VFM tokens.

\paragraph{Query, key and value.} Tab.~\ref{tab:qkv} decomposes the three prior methods and
\ours{}. All four start the query from the RGB image and take the values from $V_{LR}$ at its own
width. They differ in how the key is formed and in whether a pretrained network reaches the query.

All four pool the key to the low-resolution grid, but from different sources. JAFAR and AnyUp pool
a second encoder run over the image, then inject the target VFM into it: JAFAR modulates
the branch by $V_{LR}$ through a Spatial Feature Transform, and AnyUp concatenates $V_{LR}$ after a
learned unification that marginalizes the channel axis. NAF and \ours{} pool the query itself, so
key and query stay in one space. NAF reads $V_{LR}$ only for its spatial shape, leaving no
pretrained network in its guidance path: $1{\times}1$ pixel and $3{\times}3$ contextual
convolutions over raw pixels, trained from scratch. Where semantics enter at all, they enter on the key, from the VFM being upsampled. That coupling
shapes JAFAR's key projection to the width $d$ and costs it single-checkpoint transfer. AnyUp
escapes the cost by marginalizing the channel axis, which keeps transfer but discards width
information. \ours{} draws semantics from a third source: a frozen pretrained encoder independent of both the
image branch and the target VFM. It conditions the query through cross-attention, and the key is
pooled from that same encoding. Semantics reach both operands, and $V_{LR}$ enters only as values.

\begin{table*}
\centering

\small\setlength{\tabcolsep}{2pt}
\begin{tabular}{@{}l ll l cc ll@{}}
\toprule
& \multicolumn{2}{c}{\textit{Source}} & & \multicolumn{2}{c}{\textit{Semantics}} & & \\
\cmidrule(lr){2-3}\cmidrule(lr){5-6}
Method & Query & Key & Attention & Q & K & Origin & Mechanism \\
\midrule
JAFAR   & RGB image & Pooled branch, mod.\ by $V_{LR}$ & Global & \xmark & \cmark
        & Target VFM & SFT on the key \\
AnyUp   & RGB image & Pooled branch $\oplus$ $V_{LR}$ & Windowed & \xmark & \cmark
        & Target VFM & Concatenation \\
NAF     & RGB image & Pooled query & Neighborhood & \xmark & \xmark
        & None & None \\
\rowcolor{blue!8}
\ours{} & RGB image $+$ Sem.\ Encoder & Pooled query & Neighborhood & \cmark & \cmark
        & Separate frozen encoder & Cross-attention \\
\bottomrule
\end{tabular}
\caption[Attention operands]{\textbf{How each method builds its attention operands.} $V_{LR}$ is
the low-resolution feature map of the frozen target VFM. \emph{Source} gives the tensor each
operand is computed from, and \emph{Neighborhood} is neighborhood attention~\cite{natten}.
\emph{Semantics} marks whether a pretrained network reaches the query (Q) or the key (K), where it
originates and how it is injected; \cmark{} yes, \xmark{} no. Our row is shaded.}
\label{tab:qkv}
\end{table*}

\section{Quantitative Results}
\label{sec:quant}
This section reports metrics, configurations, and additional ablations that complement the main paper results.

\subsection{Training and Evaluation}

\subsubsection{PixelUp Training}
\label{sec:train}
\ours{} is trained once, without labels, and then frozen. A frozen DINOv3-L~\cite{dinov3} encoder acts as the teacher, and the objective is the mean squared
error to its features. Each step resizes the training image to a random square of $480$, $640$ or
$800$ pixels and forms the low-resolution input by downsampling it by a factor of $0.3$, $0.5$ or
$0.7$. Size and factor resample at every step, so no fixed upsampling factor is ever trained and
one checkpoint serves arbitrary scales at test time. Tab.~\ref{tab:train} lists the full
configuration. All images and pretrained weights are public and used under their respective licenses for
research: SA-1B, COCO and, for Sec.~\ref{sec:gen-data}, ImageNet-1k for training, and Pascal
VOC~\cite{voc}, Cityscapes~\cite{cityscapes},
ADE20K~\cite{ade20k}, NYUv2~\cite{nyuv2} and KITTI~\cite{kitti} for evaluation. Every frozen
encoder is used as released.

\begin{table}
\centering

\small\setlength{\tabcolsep}{4pt}
\begin{tabular}{@{}ll@{}}
\toprule
Teacher              & DINOv3-L, frozen \\
Objective            & mean squared error to teacher features \\
Images               & $134{,}473$, unlabeled \\
Image sizes          & $480$, $640$ or $800$ square, per step \\
Reduction factors    & $0.3$, $0.5$ or $0.7$, per step \\
Optimizer            & AdamW, lr $2{\times}10^{-4}$, decay $0.05$ \\
Schedule             & $1{,}000$-step warm-up, then cosine to $1\%$ \\
Steps / batch size   & $50{,}000$ / $2$ \\
Gradient clipping    & $1.0$ \\
Precision            & bfloat16 \\
Encoder width        & $256$ \\
Attention heads      & $4$ encoder, $4$ decoder, width $64$ \\
Neighborhood         & $9{\times}9$ window \\
Semantic Encoder     & DINOv3 ConvNeXt-S, frozen, $49.45$M \\
Trainable weights    & $6.27$M \\
Hardware             & one L40S ($48$\,GB) \\
\bottomrule
\end{tabular}
\caption[PixelUp training configuration]{\textbf{Training configuration.} The single label-free
run that produces the deployed \ours{} checkpoint. The $134{,}473$ images mix SA-1B and COCO.
At the deployed width of $256$, the $64$-dimensional heads give $4/4$ encoder and decoder heads;
six heads are not possible, since $256$ is not divisible by six.}
\label{tab:train}
\end{table}

\subsubsection{Experimental Configuration}
\label{sec:config}
The evaluation protocols run on top of the frozen checkpoint above.

\paragraph{Baseline selection.} The comparison covers every upsampler that runs under these
protocols without retraining once per backbone. NAF and AnyUp deploy a single checkpoint across
encoders, as \ours{} does, and nearest, bilinear and JBU need no weights at all. JAFAR is included
on the backbones for which public checkpoints exist. We could not evaluate
FeatSharp~\cite{ranzingerfeatsharp}, as no trained checkpoints are publicly available.

\paragraph{Linear probing.} The backbone and \ours{} both stay frozen, and we train one
$1{\times}1$ convolutional classifier on the upsampled features. Following \cite{naf}, images are resized and center-cropped to $448{\times}448$ and
scored there. Training uses AdamW at weight decay $10^{-5}$, batch size two, a cosine schedule and
full precision, for twenty epochs on VOC~\cite{voc}, Cityscapes~\cite{cityscapes} and
NYUv2~\cite{nyuv2}, and five on ADE20K~\cite{ade20k}. The learning rate is $5{\times}10^{-4}$
everywhere except Cityscapes, which uses $10^{-4}$. We report the last epoch and select no
checkpoint. Segmentation mIoU comes from a confusion matrix pooled over the validation set and
averaged over classes with equal weight.

\paragraph{Ablation protocol.} The architecture ablations in the main paper vary one design choice
at a time around a common anchor of $384$ encoder channels, with the training setup of
Tab.~\ref{tab:train} otherwise unchanged. They probe Cityscapes at the same $5{\times}10^{-4}$
rate used for VOC, not the $10^{-4}$ used above.

\paragraph{Efficiency measurement.} The efficiency figures in the main paper are measured end to
end and already include the frozen Semantic Encoder. Even at $55.72$M total parameters against
NAF's $0.66$M, \ours{} is $1.66\times$ faster ($32.3$ vs.\ $19.5$ FPS) with $1.58\times$ lower
peak memory ($1.77$ vs.\ $2.79$~GB) on DINOv3-B at $448^2$.

\paragraph{Unsupervised segmentation.} DiffCut~\cite{diffcut} groups pixels by recursive
normalized cuts over diffusion features and trains nothing. We read features from the last
self-attention layer of the SSD-1B~\cite{ssd1b} denoiser after a single forward pass at diffusion
timestep $50$. Images enter at $1024{\times}1024$,
which places the features on a $32{\times}32$ grid. Cuts use a bipartition threshold of $0.5$ and
an affinity exponent of $10$, and we refine the resulting masks with pixel-adaptive refinement module as the original method does. \ours{} replaces exactly one operation,
the bilinear interpolation that lifts the $32{\times}32$ features to the clustering grid, so
bilinear interpolation is an exact control. The default grid is $128{\times}128$. Scoring matches clusters to ground-truth classes
by Hungarian assignment per image, then pools the per-class counts across the dataset before
averaging IoU over classes.

\paragraph{Open-vocabulary segmentation.} RADSeg~\cite{radseg} labels pixels by comparing image
features against text embeddings and likewise trains nothing. A frozen C-RADIOv3-B~\cite{radio}
encoder produces the features, and its SigLIP2~\cite{siglip2} adapter maps them into the text
embedding space, each class embedding averaging eighty prompt templates. We run the encoder once
over the whole image instead of tiling it into overlapping windows, and \ours{} then lifts the
language-aligned feature map from the patch grid to the full input resolution. Class
scores follow from the lifted features, and the no-upsampler control instead enlarges the
low-resolution scores. We score with mmsegmentation's mIoU, excluding background throughout, following the official protocol.

\subsection{Seed Variance}
\label{sec:seeds}
We trained two further checkpoints from different random seeds, identical in every other respect,
and evaluated them on four benchmarks spanning all three protocols. Tab.~\ref{tab:seeds} reports a
spread between $0.02$ and $0.11$ mIoU, so differences of a few tenths in this paper are real and
differences below roughly a tenth are not. The margins over NAF are far outside that noise, more than twenty times the corresponding
deviation on Cityscapes and on KITTI.

\begin{table}
\centering

\small\setlength{\tabcolsep}{3pt}
\begin{tabular}{@{}ll rrr rr@{}}
\toprule
Benchmark & Protocol & seed $0$ & seed $1$ & seed $2$ & \textit{Mean} & \textit{Std} \\
\midrule
Cityscapes & probing      & 68.16 & 68.03 & 68.01 & 68.07 & 0.08 \\
KITTI      & open-vocab.  & 47.04 & 47.02 & 47.22 & 47.09 & 0.11 \\
ADE20K     & open-vocab.  & 30.17 & 30.27 & 30.22 & 30.22 & 0.05 \\
VOC20      & unsup.       & 65.96 & 65.99 & 65.97 & 65.97 & 0.02 \\
\bottomrule
\end{tabular}
\caption[Seed variance]{\textbf{Variance across three training seeds.} All values are
mIoU$\uparrow$. Only the random seed changes. The deployed checkpoint is seed $0$.}
\label{tab:seeds}
\end{table}

\subsection{Semantic Segmentation}
Comparing mIoU against pixel accuracy localizes where \ours{} gains on Cityscapes. The main
table's mIoU weights every class equally, while the pixel accuracy (aAcc) added by
Tab.~\ref{tab:secondary-seg} is dominated by whichever classes cover the most pixels. \ours{}
improves mIoU over NAF by $+3.20$ ($68.16$ against $64.96$) but aAcc by only $+0.69$ ($94.97$
against $94.28$). That pattern places the gain in classes occupying few pixels: the thin and
fine-grained categories and the boundaries between them, not the large uniform regions that
dominate the pixel count. Sec.~\ref{sec:boundary} measures that directly and reaches the same
conclusion.
\begin{table}
\centering

\small\setlength{\tabcolsep}{4pt}
\begin{tabular}{@{}l cc cc cc@{}}
\toprule
& \multicolumn{2}{c}{VOC} & \multicolumn{2}{c}{Cityscapes} & \multicolumn{2}{c}{ADE20K} \\
\cmidrule(lr){2-3}\cmidrule(lr){4-5}\cmidrule(lr){6-7}
Upsampler & mIoU & aAcc & mIoU & aAcc & mIoU & aAcc \\
\midrule
Nearest  & 84.96 & 96.33 & 58.79 & 92.01 & 43.58 & 76.11 \\
Bilinear & 86.88 & 96.96 & 63.01 & 93.63 & 45.53 & 77.37 \\
JBU      & 87.14 & 97.03 & 63.83 & 93.87 & 45.75 & 77.52 \\
AnyUp    & 86.58 & 96.86 & 60.31 & 93.20 & 45.19 & 77.26 \\
NAF      & 87.81 & 97.19 & 64.96 & 94.28 & 46.50 & 77.87 \\
\rowcolor{blue!8}
\textbf{\ours{}} & \textbf{88.71} & \textbf{97.41} & \textbf{68.16} & \textbf{94.97} & \textbf{47.27} & \textbf{78.15} \\
\midrule
$\Delta_{\text{NAF}}$ & \textcolor{green!45!black}{$+$0.90} & \textcolor{green!45!black}{$+$0.22} & \textcolor{green!45!black}{$+$3.20} & \textcolor{green!45!black}{$+$0.69} & \textcolor{green!45!black}{$+$0.77} & \textcolor{green!45!black}{$+$0.28} \\
\bottomrule
\end{tabular}
\caption[Secondary segmentation metric]{\textbf{Pixel accuracy alongside mIoU.} DINOv3-B, linear
probing at $448^2$; only the upsampler changes down the rows, each with its own trained probe.
aAcc is overall pixel accuracy, the fraction of labeled pixels classified correctly; both metrics
are percentages (mIoU$\uparrow$, aAcc$\uparrow$). $\Delta_{\text{NAF}}$ is \ours{} minus NAF.
\textbf{Bold} best per column; our row is shaded.}
\label{tab:secondary-seg}
\end{table}

Per-class IoU highlights the gains of \ours{} over NAF for fine-grained categories. Tab.~\ref{tab:perclass} decomposes the Cityscapes
column above using the classifiers that produced it, so the nineteen values average to the
reported mIoU, not the per-image average. The five classes covering $91\%$ of the pixels gain $1.18$ over NAF. The remaining
fourteen, covering $9\%$, gain $3.91$. The largest gains fall on thin and boundary-dominated
classes: traffic light $+7.3$, traffic sign $+5.8$, bicycle and rider $+5.3$, person $+4.7$,
sidewalk $+4.3$ and pole $+4.2$. Area alone does not predict this: bus, train and truck are just
as rare but compact. What tracks the gain is the share of a class's pixels lying near its own
boundary. Sky is the only class that loses marginally, and it is almost entirely interior.

\begin{table}
\centering

\small\setlength{\tabcolsep}{4pt}
\begin{tabular}{@{}l r rr r@{}}
\toprule
Class & \% px & NAF & \ours{} & $\Delta_{\text{NAF}}$ \\
\midrule
road          & 43.06 & 97.5 & 98.1 & \textcolor{green!45!black}{$+$0.6} \\
building      & 18.94 & 88.1 & 89.3 & \textcolor{green!45!black}{$+$1.2} \\
vegetation    & 18.49 & 89.4 & 90.2 & \textcolor{green!45!black}{$+$0.8} \\
sky           &  5.39 & 94.2 & 93.8 & \textcolor{red!65!black}{$-$0.4} \\
car           &  5.03 & 85.2 & 88.9 & \textcolor{green!45!black}{$+$3.7} \\
sidewalk      &  2.73 & 67.1 & 71.4 & \textcolor{green!45!black}{$+$4.3} \\
person        &  1.41 & 68.5 & 73.2 & \textcolor{green!45!black}{$+$4.7} \\
pole          &  1.27 & 27.6 & 31.8 & \textcolor{green!45!black}{$+$4.2} \\
traffic sign  &  0.59 & 53.0 & 58.8 & \textcolor{green!45!black}{$+$5.8} \\
bicycle       &  0.53 & 57.0 & 62.3 & \textcolor{green!45!black}{$+$5.3} \\
bus           &  0.45 & 82.8 & 85.5 & \textcolor{green!45!black}{$+$2.7} \\
wall          &  0.44 & 47.4 & 49.5 & \textcolor{green!45!black}{$+$2.1} \\
terrain       &  0.38 & 46.8 & 48.9 & \textcolor{green!45!black}{$+$2.1} \\
fence         &  0.35 & 41.5 & 43.9 & \textcolor{green!45!black}{$+$2.4} \\
truck         &  0.33 & 76.7 & 79.4 & \textcolor{green!45!black}{$+$2.7} \\
rider         &  0.23 & 43.2 & 48.5 & \textcolor{green!45!black}{$+$5.3} \\
traffic light &  0.18 & 43.9 & 51.2 & \textcolor{green!45!black}{$+$7.3} \\
train         &  0.11 & 75.9 & 78.2 & \textcolor{green!45!black}{$+$2.3} \\
motorcycle    &  0.08 & 48.6 & 52.2 & \textcolor{green!45!black}{$+$3.6} \\
\midrule
mean          &       & 64.96 & 68.16 & \textcolor{green!45!black}{$+$3.20} \\
\bottomrule
\end{tabular}
\caption[Per-class Cityscapes IoU]{\textbf{Per-class IoU on Cityscapes.} DINOv3-B, linear probing,
all nineteen classes ordered by pixel share; IoU$\uparrow$ in percent. \% px is the share of
labeled pixels pooled over the validation split. $\Delta_{\text{NAF}}$ is \ours{} minus NAF (\textcolor{green!45!black}{green} gain,
\textcolor{red!65!black}{red} loss), and \textit{mean} is the unweighted average over the nineteen
classes.}
\label{tab:perclass}
\end{table}

\subsection{Monocular Depth}
On NYUv2 the metrics separate the methods far less than the segmentation results do. In
Tab.~\ref{tab:secondary-depth}, $\delta_2$ and $\delta_3$ match for both
\ours{} and NAF ($98.9$ and $99.9$), so neither discriminates. $\delta_1$ and RMSE carry the
whole signal: $\delta_1$ separates the two methods by $0.2$ points ($88.5$ against $88.3$) and
RMSE by two millimeters ($0.399$ against $0.401$~m). RMSE agrees with $\delta_1$ on the ordering.
\begin{table}
\centering

\small\setlength{\tabcolsep}{5pt}
\begin{tabular}{@{}l cccc@{}}
\toprule
Upsampler & RMSE$\downarrow$ & $\delta_1\uparrow$ & $\delta_2\uparrow$ & $\delta_3\uparrow$ \\
\midrule
Nearest  & 0.424 & 87.1 & 98.5 & 99.8 \\
Bilinear & 0.407 & 87.9 & 98.8 & 99.9 \\
JBU      & 0.404 & 88.1 & 98.8 & 99.9 \\
AnyUp    & 0.407 & 88.2 & 98.8 & 99.8 \\
NAF      & 0.401 & 88.3 & 98.9 & 99.9 \\
\rowcolor{blue!8}
\textbf{\ours{}} & \textbf{0.399} & \textbf{88.5} & 98.9 & 99.9 \\
\midrule
$\Delta_{\text{NAF}}$ & \textcolor{green!45!black}{$-$0.002} & \textcolor{green!45!black}{$+$0.2} & 0.0 & 0.0 \\
\bottomrule
\end{tabular}
\caption[Secondary depth metrics]{\textbf{Secondary depth metrics.} NYUv2, DINOv3-B, linear
probing; RMSE is in meters; $\delta_k$ is the percentage of pixels whose predicted-to-true depth ratio lies within
$1.25^{k}$. $\Delta_{\text{NAF}}$ is \ours{} minus NAF, so the negative RMSE entry is a gain.
\textbf{Bold} best per column, left unset where the best value is tied; our row is shaded.}
\label{tab:secondary-depth}
\end{table}

\subsection{Unsupervised Segmentation}
Tab.~\ref{tab:diffcut-res} sweeps the output mask grid, the one quantity an
upsampler controls. The parameter-free baselines get monotonically \emph{worse} in every column as
the grid grows (nearest neighbor on ADE20K: $44.44 \to 43.07 \to 42.12$), because they interpolate
rather than reconstruct. The learned baselines stop running altogether: AnyUp exhausts memory at
the $512$ grid and NAF at $1024$. \ours{} gains through the $512$ grid and holds at $1024$
($44.55 \to 45.09 \to 44.99$), so its margin over the second-best runnable method widens from
$+0.36$ mean at $128$ to $+0.91$ at $512$ and $+2.85$ at $1024$, where bilinear is the strongest
surviving baseline.

\begin{table}
\centering

\small\setlength{\tabcolsep}{4pt}
\begin{tabular}{@{}l ccc c@{}}
\toprule
Method & ADE20K & Cityscapes & KITTI & \textit{Avg} \\
\midrule
\multicolumn{5}{@{}l}{\textit{mask grid }$128$}\\
Nearest  & 44.44 & 30.47 & \underline{33.46} & 36.12 \\
Bilinear & 44.37 & 30.50 & 33.44 & 36.10 \\
AnyUp    & 44.38 & 30.95 & 33.09 & 36.14 \\
NAF      & \underline{44.49} & \underline{31.00} & 33.40 & \underline{36.30} \\
\rowcolor{blue!8}
\textbf{\ours{}} & \textbf{44.55} & \textbf{31.45} & \textbf{33.97} & \textbf{36.66} \\
$\Delta_{\text{prev}}$ & \textcolor{green!45!black}{$+$0.06} & \textcolor{green!45!black}{$+$0.45} & \textcolor{green!45!black}{$+$0.51} & \textcolor{green!45!black}{$+$0.36} \\
\midrule
\multicolumn{5}{@{}l}{\textit{mask grid }$512$}\\
Nearest  & 43.07 & 29.35 & 31.96 & 34.79 \\
Bilinear & 43.55 & 29.68 & 32.52 & 35.25 \\
AnyUp    & \textcolor{red!65!black}{OOM} & \textcolor{red!65!black}{OOM} & \textcolor{red!65!black}{OOM} & -- \\
NAF      & \underline{44.84} & \underline{31.25} & \underline{33.13} & \underline{36.41} \\
\rowcolor{blue!8}
\textbf{\ours{}} & \textbf{45.09} & \textbf{32.51} & \textbf{34.36} & \textbf{37.32} \\
$\Delta_{\text{prev}}$ & \textcolor{green!45!black}{$+$0.25} & \textcolor{green!45!black}{$+$1.26} & \textcolor{green!45!black}{$+$1.23} & \textcolor{green!45!black}{$+$0.91} \\
\midrule
\multicolumn{5}{@{}l}{\textit{mask grid }$1024$}\\
Nearest  & 42.12 & 28.30 & 31.46 & 33.96 \\
Bilinear & \underline{42.80} & \underline{28.75} & \underline{31.80} & \underline{34.45} \\
AnyUp    & \textcolor{red!65!black}{OOM} & \textcolor{red!65!black}{OOM} & \textcolor{red!65!black}{OOM} & -- \\
NAF      & \textcolor{red!65!black}{OOM} & \textcolor{red!65!black}{OOM} & \textcolor{red!65!black}{OOM} & -- \\
\rowcolor{blue!8}
\textbf{\ours{}} & \textbf{44.99} & \textbf{32.48} & \textbf{34.44} & \textbf{37.30} \\
$\Delta_{\text{prev}}$ & \textcolor{green!45!black}{$+$2.19} & \textcolor{green!45!black}{$+$3.73} & \textcolor{green!45!black}{$+$2.64} & \textcolor{green!45!black}{$+$2.85} \\
\bottomrule
\end{tabular}
\caption[Unsupervised segmentation across output resolutions]{\textbf{Unsupervised segmentation
across output resolutions} (DiffCut, SSD-1B; mIoU$\uparrow$). Each block fixes the mask grid, the
clustering resolution the $32{\times}32$ features are lifted to, and varies only the operation
that lifts them. \textit{Avg} averages the three benchmarks shown; OOM marks a run that exhausts
GPU memory. $\Delta_{\text{prev}}$ is \ours{} minus the best other method per column in its block.
\textbf{Bold} best, \underline{underline} second among runnable methods; our rows are shaded.}
\label{tab:diffcut-res}
\end{table}

\begin{table*}
\centering

\small\setlength{\tabcolsep}{5pt}
\begin{tabular}{@{}l cc cc cc cc cc@{}}
\toprule
& \multicolumn{2}{c}{ADE20K} & \multicolumn{2}{c}{Cityscapes} & \multicolumn{2}{c}{KITTI}
& \multicolumn{2}{c}{VOC20} & \multicolumn{2}{c}{Mean} \\
\cmidrule(lr){2-3}\cmidrule(lr){4-5}\cmidrule(lr){6-7}\cmidrule(lr){8-9}\cmidrule(lr){10-11}
Inference & mIoU & mAcc & mIoU & mAcc & mIoU & mAcc & mIoU & mAcc & mIoU & mAcc \\
\midrule
Tiled windows        & 28.96 & 55.03 & 45.41 & \textbf{67.49} & 41.91 & 69.20 & 89.28 & 94.56 & 51.39 & 71.57 \\
\midrule
\multicolumn{11}{@{}l}{\textit{Single whole-image pass}}\\
No upsampler         & 28.23 & 55.64 & 42.05 & 65.30 & 41.23 & 69.36 & 87.88 & 93.66 & 49.85 & 70.99 \\
NAF                  & \textbf{30.56} & 55.89 & 43.42 & 60.92 & 44.37 & 66.23 & \textbf{89.49} & \textbf{94.60} & 51.96 & 69.41 \\
\rowcolor{blue!8}
\ours{}              & 30.17 & \textbf{56.28} & \textbf{46.71} & 66.34 & \textbf{47.04} & \textbf{70.29} & 89.25 & 94.37 & \textbf{53.29} & \textbf{71.82} \\
\midrule
$\Delta_{\text{tiled}}$
 & \textcolor{green!45!black}{$+$1.21} & \textcolor{green!45!black}{$+$1.25}
 & \textcolor{green!45!black}{$+$1.30} & \textcolor{red!65!black}{$-$1.15}
 & \textcolor{green!45!black}{$+$5.13} & \textcolor{green!45!black}{$+$1.09}
 & \textcolor{red!65!black}{$-$0.03} & \textcolor{red!65!black}{$-$0.19}
 & \textcolor{green!45!black}{$+$1.90} & \textcolor{green!45!black}{$+$0.25} \\
\bottomrule
\end{tabular}
\caption[Tiled inference against one pass]{\textbf{Tiled inference against a single whole-image
pass.} RADSeg with a frozen C-RADIOv3-B encoder and its SigLIP2 adapter at the resolutions of
Sec.~\ref{sec:config}. mIoU$\uparrow$ and
mAcc$\uparrow$ are percentages. $\Delta_{\text{tiled}}$ is
\ours{} minus the tiled row. \textbf{Bold} best per column; our row is shaded.}
\label{tab:tiled}
\end{table*}

\subsection{Open-Vocabulary Segmentation}
Tab.~\ref{tab:tiled} compares RADSeg's original tiling, which encodes overlapping windows and blends them, against the single
whole-image pass of Sec.~\ref{sec:config}. Averaged over the four benchmarks, \ours{} beats the tiled protocol by $1.90$ mIoU while holding
recall level, $71.82$ against $71.57$ mAcc. NAF gains $0.57$ and loses recall, falling to $69.41$.
Cityscapes separates the two most clearly: the lift has to recover $3.36$ mIoU to match tiling,
NAF recovers $1.37$ of that, and \ours{} recovers all of it and $1.30$ more. Discarding the
windowing heuristic is safe only when the upsampler is accurate enough to replace it.

\FloatBarrier
\section{Boundary-Aware Evaluation}
\label{sec:boundary}
Whole-image metrics are dominated by region interiors, where the upsamplers differ least, so they
cannot isolate the performance on fine-grained details. This section restricts both dense tasks to the pixels
near an object boundary. Tab.~\ref{tab:boundary} does so two ways, by trimap band and by contour
F-score \cite{cheng2021boundary}. The margin over NAF \emph{widens} as the evaluation tightens onto boundaries: $+3.20$ mIoU over
the whole image, $+3.67$ in a $16$-pixel band and $+4.29$ in an $8$-pixel band, holding at $+4.20$
in a $4$-pixel band. Boundary F-score agrees
independently ($68.35$ against NAF's $66.42$ at $t{=}2$), so the gain is contour placement and not
merely cleaner region labeling. Bilinear makes the contour metric legible: it trails \ours{} by
$5.15$ mIoU over the whole image ($63.01$) but by $16.10$ points of F$@2$ ($52.25$).
Tab.~\ref{tab:perclass} reached the same conclusion through the classes that gain rather than the
pixels that do.
\begin{table}
\centering
{\small
\setlength{\tabcolsep}{4pt}
\renewcommand{\arraystretch}{1.1}
\begin{tabular}{@{}l cccc cc@{}}
\toprule
& \multicolumn{4}{c}{\textbf{\textit{Trimap mIoU}}$\uparrow$} & \multicolumn{2}{c}{\textbf{\textit{Boundary F}}$\uparrow$} \\
\cmidrule(lr){2-5}\cmidrule(lr){6-7}
Method & all & @16 & @8 & @4 & @2 & @4 \\
\midrule
Bilinear & 63.01 & 59.42 & 53.76 & 45.65 & 52.25 & 70.24 \\
AnyUp    & 60.31 & 56.32 & 50.89 & 44.66 & \underline{66.51} & \underline{77.21} \\
NAF      & \underline{64.96} & \underline{61.62} & \underline{56.48} & \underline{49.84} & 66.42 & 76.91 \\
\rowcolor{blue!8}
\textbf{\ours{}} & \textbf{68.16} & \textbf{65.29} & \textbf{60.77} & \textbf{54.04} & \textbf{68.35} & \textbf{78.92} \\
\midrule
\textbf{\textit{$\Delta_{\text{NAF}}$}}
 & \textbf{\textcolor{green!45!black}{$+$3.20}}
 & \textbf{\textcolor{green!45!black}{$+$3.67}}
 & \textbf{\textcolor{green!45!black}{$+$4.29}}
 & \textbf{\textcolor{green!45!black}{$+$4.20}}
 & \textbf{\textcolor{green!45!black}{$+$1.93}}
 & \textbf{\textcolor{green!45!black}{$+$2.01}} \\
\bottomrule
\end{tabular}
}
\caption[Boundary-aware segmentation]{\textbf{Boundary-aware segmentation.} Cityscapes, DINOv3-B,
reusing the frozen backbone, upsampler and probe of Tab.~\ref{tab:secondary-seg}; only the metric
changes across columns. \textit{Trimap mIoU}@$w$ scores only pixels within $w$ of a ground-truth
class boundary and \textit{all} the whole image; \textit{Boundary F}@$t$ is the contour F-score at
a tolerance of $t$ pixels. Both are percentages. $\Delta_{\text{NAF}}$ is \ours{} minus NAF.
\textbf{Bold} best, \underline{underline} second per column; our row is shaded.}
\label{tab:boundary}
\end{table}

The same restriction applies to depth, where the whole-image result is the closest to a tie in the
paper. Tab.~\ref{tab:boundary-depth} scores $\delta_1$ inside trimap bands around ground-truth
depth discontinuities and adds RMSE restricted to silhouette pixels, and the pattern of
Tab.~\ref{tab:boundary} repeats. The margin over NAF widens from $+0.18$ over the whole image to
$+0.27$ in an eight-pixel band, and \ours{} is the only method that scores higher inside every
band than over the whole image, by $0.07$, $0.06$ and $0.05$. NAF gains $0.01$ at sixteen pixels
and then loses $0.04$ by four, and AnyUp falls monotonically to $0.12$ below its own whole-image
value. 

\begin{table}
\centering
{\small
\setlength{\tabcolsep}{4pt}
\renewcommand{\arraystretch}{1.1}
\begin{tabular}{@{}l cccc cc@{}}
\toprule
& \multicolumn{4}{c}{\textbf{\textit{Trimap} $\delta_1$}$\uparrow$}
& \multicolumn{2}{c}{\textbf{\textit{Silhouette RMSE}}$\downarrow$} \\
\cmidrule(lr){2-5}\cmidrule(lr){6-7}
Method & all & @16 & @8 & @4 & all & @4 \\
\midrule
AnyUp & 88.16 & 88.14 & 88.07 & 88.04 & 0.407 & 0.404 \\
NAF   & \underline{88.32} & \underline{88.33} & \underline{88.29} & \underline{88.28} & \underline{0.401} & \underline{0.396} \\
\rowcolor{blue!8}
\textbf{\ours{}} & \textbf{88.50} & \textbf{88.57} & \textbf{88.56} & \textbf{88.55}
                 & \textbf{0.399} & \textbf{0.395} \\
\midrule
\textbf{\textit{$\Delta_{\text{NAF}}$}}
 & \textbf{\textcolor{green!45!black}{$+$0.18}} & \textbf{\textcolor{green!45!black}{$+$0.24}}
 & \textbf{\textcolor{green!45!black}{$+$0.27}} & \textbf{\textcolor{green!45!black}{$+$0.27}}
 & \textbf{\textcolor{green!45!black}{$-$0.002}} & \textbf{\textcolor{green!45!black}{$-$0.001}} \\
\bottomrule
\end{tabular}
}
\caption[Boundary-aware depth]{\textbf{Boundary-aware depth.} NYUv2, DINOv3-B, reusing the frozen
backbone, upsampler and head of Tab.~\ref{tab:secondary-depth}; only the metric changes.
\textit{Trimap} $\delta_1$@$w$ scores only pixels within $w$ of a ground-truth depth discontinuity
and \textit{all} the whole image, in percent; \textit{Silhouette RMSE} restricts RMSE to those
pixels, in meters. $\Delta_{\text{NAF}}$ is \ours{} minus NAF, so the negative RMSE entries are
gains. \textbf{Bold} best, \underline{underline} second per column; our row is shaded.}
\label{tab:boundary-depth}
\end{table}

\FloatBarrier
\section{The Semantic Encoder}
\label{sec:sem-enc}
\ours{} builds the semantics of its query from a frozen Semantic Encoder. This section isolates
that component and asks two questions: \emph{what} the encoder contributes
(Sec.~\ref{sec:guid-vs-cap}), by varying it from the deployed DINOv3 ConvNeXt-S down to no encoder
at all, and \emph{how} its features should enter the query (Sec.~\ref{sec:fusion}).

\begin{table*}
\centering
\small\setlength{\tabcolsep}{7pt}
\begin{tabular}{@{}l rr cc cc cc@{}}
\toprule
& \multicolumn{2}{c}{\textit{Params (M)}} & \multicolumn{2}{c}{\textit{Linear probing}} & \multicolumn{2}{c}{\textit{Open-vocab.}} & & \\
\cmidrule(lr){2-3}\cmidrule(lr){4-5}\cmidrule(lr){6-7}
Semantic Encoder & Frozen & Train. & VOC & CS & KITTI & ADE & \textit{Mean} & $\Delta_{\text{NAF}}$ \\
\midrule
\rowcolor{blue!8}
DINOv3 ConvNeXt-S \textit{(deployed)} & 49.45 & 6.27 & \textbf{88.71} & \textbf{68.16} & \textbf{47.04} & 30.17 & \textbf{58.52} & \textbf{\textcolor{green!45!black}{$+$1.59}} \\
DINOv3 ConvNeXt-T & 27.82 & 6.27 & \textbf{88.71} & 68.13 & 46.87 & 30.20 & \underline{58.48} & \underline{\textcolor{green!45!black}{$+$1.55}} \\
ConvNeXt-V2-Atto & 3.39 & 3.94 & 88.52 & 67.39 & 46.23 & 30.21 & 58.09 & \textcolor{green!45!black}{$+$1.16} \\
\midrule
\textit{NAF (prior SOTA)} & -- & 0.66 & 87.81 & 64.96 & 44.37 & \textbf{30.56} & 56.92 & -- \\
None (Pixel Encoder only) & -- & 0.50 & 87.21 & 62.85 & 44.42 & 30.24 & 56.18 & \textcolor{red!65!black}{$-$0.74} \\
\midrule
\multicolumn{9}{@{}l}{\textit{ConvNeXt-S architecture without its pretraining}}\\
Trained from scratch, $50$k steps  & -- & 55.73 & 86.53 & 61.57 & 40.59 & 29.89 & 54.64 & \textcolor{red!65!black}{$-$2.28} \\
Trained from scratch, $150$k steps & -- & 55.73 & 86.47 & 58.53 & 38.08 & 29.22 & 53.07 & \textcolor{red!65!black}{$-$3.85} \\
Random initialization, frozen & 49.45 & 6.27 & 86.55 & 61.46 & 40.54 & 29.84 & 54.60 & \textcolor{red!65!black}{$-$2.33} \\
Permuted pretrained weights, frozen & 49.45 & 6.27 & 84.36 & 56.11 & 39.45 & 29.38 & 52.33 & \textcolor{red!65!black}{$-$4.60} \\
\bottomrule
\end{tabular}
\caption[Semantic Encoder ablation]{\textbf{Semantic Encoder ablation.} Each row retrains
\ours{} with a different Semantic Encoder; all other settings follow Tab.~\ref{tab:train}. All
results are mIoU$\uparrow$; $\Delta_{\text{NAF}}$ is the mean relative to NAF. The bottom block
keeps the ConvNeXt-S architecture and removes only its pretraining; \emph{permuted} shuffles the
entries of each pretrained tensor, preserving the weight statistics while destroying the learned
structure.}
\label{tab:gen-guide}
\end{table*}

\subsection{Guidance versus Capacity}
\label{sec:guid-vs-cap}
Tab.~\ref{tab:gen-guide} retrains \ours{} with different Semantic Encoders while keeping the
teacher and all other settings fixed.

\paragraph{Removing the encoder.} Without the Semantic Encoder, the mean over the four benchmarks
of Tab.~\ref{tab:gen-guide} (VOC, Cityscapes, KITTI, ADE20K) drops by $2.34$, more than
\ours{}'s entire $1.59$ margin over NAF, leaving the model $0.74$ \emph{below}
the prior state of the art. The loss concentrates where fine
structure is measured. Cityscapes falls $5.31$ and KITTI $2.62$, while ADE20K, dominated by large
regions at $448^2$, is unchanged ($+0.07$). KITTI and ADE20K share the same open-vocabulary
protocol and frozen backbone, so this split reflects image content rather than evaluation. That is
the signature of a semantic prior, not of added capacity. This variant, however, also removes
$92\%$ of the trainable parameters ($6.27$M$\,\rightarrow\,0.50$M), conflating guidance with
capacity. The two smaller encoders disentangle them.

\paragraph{Capacity.} DINOv3 ConvNeXt-T reduces the frozen encoder by $44\%$
($49.45\rightarrow27.82$M) at an identical trainable budget. The mean changes by $0.04$, within
the $0.08$ seed deviation of Tab.~\ref{tab:seeds}. The encoder's contribution does not depend on
its size.

\paragraph{Pretraining lineage.} ConvNeXt-V2-Atto~\cite{convnextv2} is $14.6\times$ smaller
($3.39$M), is pretrained with FCMAE on ImageNet-1k, and shares no lineage with DINOv3 or the
distillation teacher. Despite a smaller trainable budget ($3.94$M vs.\ $6.27$M), it retains
$73\%$ of the margin over NAF ($+1.16$ vs.\ $+1.59$) and $82\%$ of the gap to the encoder-free
model. Any reasonably pretrained encoder therefore suffices. The advantage is not specific to
DINOv3.

\paragraph{Pretraining.} Pretrained weights, not the multi-scale architecture, carry the gain.
The bottom block of Tab.~\ref{tab:gen-guide} keeps the ConvNeXt-S architecture and removes only
what it learned. A frozen, randomly initialized encoder scores $61.46$ on Cityscapes, $1.39$
\emph{below} removing the encoder outright. Training that encoder from scratch recovers $0.11$,
though it carries $55.73$M trainable parameters, nearly $9\times$ the deployed budget; tripling
its schedule costs a further $3.04$, as reconstruction improves while the features it yields grow
less useful downstream. Permuting each pretrained tensor preserves its mean, variance and value
distribution and destroys only its learned structure, which separates structure from scale. ~\cite{frankleearly} permute within convolutional filters early in training and lose
little accuracy; on a fully pretrained encoder the same operation is the most damaging variant we
test, $-12.05$ on Cityscapes against the deployed encoder and $-6.74$ against none. An
unpretrained pyramid therefore harms the query: image evidence alone serves it better than
uninformative multi-scale features. Removing pretrained semantics costs $2.34$ mean mIoU,
exceeding the $1.59$ margin over prior work; replacing them with unpretrained weights of the same
shape costs $6.19$; shrinking the encoder that carries them costs $0.04$.

\subsection{Fusion Operator}
\label{sec:fusion}
The semantics carry the result. The operator that injects them matters relatively less. Where changing
the encoder moves Cityscapes by $5$ to $12$ points, replacing cross-attention with concatenation
of the same semantic features moves the mean by $0.03$ at the ablation anchor of
Sec.~\ref{sec:config}, within seed deviation. To further isolate the operator, we vary the upsampling ratio while holding the encoder and all other settings fixed. Tab.~\ref{tab:ratio} fixes the output grid,
probe head, and labels at $448^2$ and varies only the backbone input ($896$, $448$, $224$),
yielding feature grids of $56$, $28$, and $14$ and upsampling ratios of $\times 8$, $\times 16$,
and $\times 32$. At the deployed $\times 16$ the operators are tied ($\Delta{=}0.08$, the seed
deviation itself). At $\times 8$ concatenation is marginally ahead. At $\times 32$
cross-attention leads by $0.67$. The mean difference grows
monotonically with the ratio: the coarse-to-fine Cross-Attention Chain earns its cost at large
upsampling factors, as its structure predicts. We therefore deploy the Cross-Attention Chain:
\ours{} targets arbitrary output resolution, and the chain's advantage grows with the ratio.

\begin{table}
\centering
{\small
\setlength{\tabcolsep}{4pt}
\renewcommand{\arraystretch}{1.1}
\begin{tabular}{@{}l cc c cc c c@{}}
\toprule
& \multicolumn{3}{c}{\textit{VOC}} & \multicolumn{3}{c}{\textit{Cityscapes}} & \\
\cmidrule(lr){2-4}\cmidrule(lr){5-7}
Ratio & cross & concat & $\Delta$ & cross & concat & $\Delta$ & \textit{Mean }$\Delta$ \\
\midrule
$\times 8$  & \textbf{88.88} & 88.84 & \textcolor{green!45!black}{$+$0.04}
            & 70.73 & \textbf{71.39} & \textcolor{red!65!black}{$-$0.66}
            & \textcolor{red!65!black}{$-$0.31} \\
\rowcolor{blue!8}
$\times 16$ & \textbf{88.75} & 88.56 & \textcolor{green!45!black}{$+$0.19}
            & 69.12 & \textbf{69.15} & \textcolor{red!65!black}{$-$0.03}
            & \textcolor{green!45!black}{$+$0.08} \\
$\times 32$ & \textbf{86.35} & 85.66 & \textcolor{green!45!black}{$+$0.69}
            & \textbf{60.71} & 60.06 & \textcolor{green!45!black}{$+$0.65}
            & \textcolor{green!45!black}{$+$0.67} \\
\bottomrule
\end{tabular}
}
\caption[Fusion operator vs.\ upsampling ratio]{\textbf{Fusion operator vs.\ upsampling ratio.}
Linear probing on DINOv3-B at $d_{\text{enc}}{=}256$ under the ablation protocol of
Sec.~\ref{sec:config}. The output grid, probe head, and labels are fixed at $448^2$; only the
backbone input varies. mIoU$\uparrow$; $\Delta$ is cross minus concat, \textbf{bold} marks the
better operator.}
\label{tab:ratio}
\end{table}

\section{Generalization}
\label{sec:gen}
The deployed checkpoint is distilled from one teacher, trained on one dataset, and zero-shot evaluated on other VFMs. This section replaces each in turn and asks what survives.

\subsection{The Distillation Teacher}
Tab.~\ref{tab:gen-teacher} retrains the deployed model
against four teachers and changes nothing else. RADIOv4 is wider than
DINOv3-L, $1152$ against $1024$, but is the weaker teacher. DINOv3-S at $384$ dimensions matches the deployed teacher: it leads on the mean over the four evaluation backbones, DINOv3-L on the mean over the four benchmarks, while RADIOv4 is
third and DINOv2-R~\cite{dinov2} last under both. Tab.~\ref{tab:gen-teacher-bb} suggests that the ordering does not reflect a family
match between teacher and evaluation backbone. The gap between the deployed teacher and RADIOv4 is
$+1.69$, $+1.48$, $+1.66$ and $+1.34$ across the four evaluation backbones, a spread of $0.35$
that does not collapse on RADIOv4's own features, where a family match would favor the
RADIOv4-distilled model. The same holds on DINOv2-R's own features: the RADIOv4-distilled model
scores $66.62$ there against the DINOv2-R-distilled model's $64.98$, so a cross-family teacher
beats the matched one on its own family's features.

Under Cityscapes probing DINOv2-R falls \emph{below}
NAF on all four backbones, by $0.54$, $1.08$, $1.33$ and $1.23$, and it also trails NAF on the
four-benchmark mean above. Hence, a poor teacher can affect the generalization of the distilled model, across both backbones and benchmarks. Also, the advantage does not follow from a larger teacher. NAF also trains its own guidance branch against
a single VFM, and its standard configuration uses DINOv3-B~\cite{naf}, one size below the DINOv3-L
deployed here. But trained from DINOv3-S, at $384$ dimensions half the width of NAF's own teacher,
\ours{} still leads NAF on all four evaluation backbones in Tab.~\ref{tab:gen-teacher-bb}, by
$3.32$, $2.81$, $1.92$ and $1.89$.

\begin{table*}
\centering

\small\setlength{\tabcolsep}{6pt}
\begin{tabular}{@{}l c cc cc cc@{}}
\toprule
& & \multicolumn{2}{c}{\textit{Linear probing}} & \multicolumn{2}{c}{\textit{Open-vocabulary}} & & \\
\cmidrule(lr){3-4}\cmidrule(lr){5-6}
Teacher & dim & VOC & Cityscapes & KITTI & ADE20K & \textit{Mean} & $\Delta_{\text{NAF}}$ \\
\midrule
\rowcolor{blue!8}
DINOv3-L \textit{(deployed)} & 1024 & 88.71 & 68.16 & \textbf{47.04} & 30.17 & \textbf{58.52} & \textbf{\textcolor{green!45!black}{$+$1.59}} \\
DINOv3-S                     & 384  & \textbf{88.83} & \textbf{68.28} & 46.39 & 30.34 & \underline{58.46} & \underline{\textcolor{green!45!black}{$+$1.54}} \\
RADIOv4                    & 1152 & 88.50 & 66.47 & 46.78 & 30.17 & 57.98 & \textcolor{green!45!black}{$+$1.06} \\
DINOv2-R                     & 768  & 87.66 & 64.42 & 45.55 & 29.82 & 56.86 & \textcolor{red!65!black}{$-$0.06} \\
\midrule
\textit{NAF (prior SOTA)}    & --   & 87.81 & 64.96 & 44.37 & \textbf{30.56} & 56.92 & -- \\
\bottomrule
\end{tabular}
\caption[Generalization across distillation teachers]{\textbf{Generalization across distillation
teachers.} Each row retrains \ours{} against a different teacher, the rest of Tab.~\ref{tab:train}
fixed. \textit{dim} is the teacher's feature width; all columns are mIoU$\uparrow$: VOC and
Cityscapes are linear probing on DINOv3-B, KITTI and ADE20K open-vocabulary transfer on
C-RADIOv3-B (Sec.~\ref{sec:config}). \textit{Mean} averages the four; $\Delta_{\text{NAF}}$ is that mean minus
NAF's, computed before rounding. \textbf{Bold} best per column,
\underline{underline} second-best mean; the deployed row is shaded.}
\label{tab:gen-teacher}
\end{table*}

\begin{table}
\centering

\small\setlength{\tabcolsep}{3pt}
\begin{tabular}{@{}l cccc@{}}
\toprule
& \multicolumn{4}{c}{\textit{evaluation backbone}} \\
\cmidrule(lr){2-5}
Teacher & DINOv3-B & RADIOv4 & DINOv2-R & Franca-B \\
\midrule
\rowcolor{blue!8}
DINOv3-L  & 68.16 & 70.19 & \textbf{68.28} & 64.35 \\
DINOv3-S  & \textbf{68.28} & \textbf{70.43} & 68.23 & \textbf{64.58} \\
RADIOv4 & 66.47 & 68.71 & 66.62 & 63.01 \\
DINOv2-R  & 64.42 & 66.54 & 64.98 & 61.46 \\
\midrule
\textit{NAF} & 64.96 & 67.62 & 66.31 & 62.69 \\
\midrule
$\Delta_{\text{NAF}}$ & \textcolor{green!45!black}{$+$3.20} & \textcolor{green!45!black}{$+$2.57} & \textcolor{green!45!black}{$+$1.97} & \textcolor{green!45!black}{$+$1.66} \\
\bottomrule
\end{tabular}
\caption[Teacher against evaluation backbone]{\textbf{Distillation teacher against evaluation
backbone.} Cityscapes linear probing (mIoU$\uparrow$). Rows vary the distillation teacher, columns
the frozen target VFM upsampled at evaluation. $\Delta_{\text{NAF}}$ is the deployed DINOv3-L row
minus NAF. \textbf{Bold} best per column; the deployed configuration is shaded.}
\label{tab:gen-teacher-bb}
\end{table}

\subsection{Training Data}
\label{sec:gen-data}
Tab.~\ref{tab:gen-data} retrains \ours{} on ImageNet-1k~\cite{imagenet} alone, an object-centric
classification corpus with no masks, instead of the deployed mix of SA-1B and COCO, and changes
nothing else. Cityscapes falls $0.53$ and VOC20 $0.12$, KITTI and ADE20K stay within their seed deviations (see Tab.~\ref{tab:seeds}). Training reads no annotation in either case, so the difference can
only come from image content: SA-1B and COCO supply dense, multi-object, high-resolution scenes
that resemble the Cityscapes distribution, whereas ImageNet images are object-centric and smaller.

\begin{table}
\centering
\small\setlength{\tabcolsep}{5pt}
\begin{tabular}{@{}l c cc c@{}}
\toprule
& \textit{Probing} & \multicolumn{2}{c}{\textit{Open-vocabulary}} & \textit{Unsup.} \\
\cmidrule(lr){2-2}\cmidrule(lr){3-4}\cmidrule(lr){5-5}
Training corpus & Cityscapes & KITTI & ADE20K & VOC20 \\
\midrule
\rowcolor{blue!8}
SA-1B $+$ COCO & \textbf{68.16} & \textbf{47.04} & 30.17 & \textbf{65.96} \\
ImageNet-1k    & 67.63 & \textbf{47.04} & 30.13 & 65.84 \\
\midrule
$\Delta$ & \textcolor{red!65!black}{$-$0.53} & 0.00
         & \textcolor{red!65!black}{$-$0.04} & \textcolor{red!65!black}{$-$0.12} \\
\midrule
\textit{NAF (prior SOTA)} & 64.96 & 44.37 & \textbf{30.56} & 65.21 \\
\bottomrule
\end{tabular}
\caption[Generalization across training corpora]{\textbf{Generalization across training datasets.}
Only the training images change; the rest of Tab.~\ref{tab:train} is unchanged. All columns are
mIoU$\uparrow$: linear probing, open-vocabulary transfer and unsupervised clustering
(\textit{Unsup.}). \textbf{Bold} best per column, with the tied best on
KITTI bolded in both rows; the deployed row is shaded.}
\label{tab:gen-data}
\end{table}

\subsection{Evaluation Protocols}
\label{sec:gen-arena}

Linear probing runs on DINOv3-B, the teacher's own family, and open-vocabulary transfer on
C-RADIOv3-B, which shares a family with the RADIOv4 teacher. Tab.~\ref{tab:gen-arena} adds two
families: SSD-1B diffusion features, which use no trained probe, and Franca-B~\cite{franca},
which shares no lineage with any teacher or Semantic Encoder here. Removing the Semantic Encoder
costs $5.31$ mIoU on DINOv3-B and $3.80$ on the neutral Franca-B, dropping the model $2.11$ and
$2.14$ below NAF. It costs $2.62$ under open-vocabulary transfer and $0.36$ on DiffCut, where all
eight rows of the column span only $0.58$ in total.

Capacity matters far less. Halving the Semantic Encoder costs at most $0.17$ on any family, and a
$384$-dimensional teacher stays within $0.25$ of the deployed $1024$-dimensional one on three of
the four. What the method needs is a pretrained encoder, not a large one. The weakest teacher is
the exception worth naming: DINOv2-R clears NAF on C-RADIO and SSD-1B but falls $0.54$ and $1.23$
below it on DINOv3-B and Franca-B, so the ordering of teachers holds across families while the
question of whether the weakest teacher clears NAF does not.
\begin{table}
\centering

\small\setlength{\tabcolsep}{2pt}
\begin{tabular}{@{}l cccc@{}}
\toprule
& \multicolumn{4}{c}{\textit{target VFM family}} \\
\cmidrule(lr){2-5}
Ablation & DINOv3-B & C-RADIO & SSD-1B & Franca-B \\
\midrule
\rowcolor{blue!8}
\ours{} \textit{(deployed)} & 68.16 & \textbf{47.04} & 33.97 & 64.35 \\
\midrule
\multicolumn{5}{@{}l}{\textit{distillation teacher}}\\
DINOv3-S      & \textbf{68.28} & 46.39 & 33.85 & \textbf{64.58} \\
RADIOv4     & 66.47 & 46.78 & 33.65 & 63.01 \\
DINOv2-R      & 64.42 & 45.55 & 33.48 & 61.46 \\
\midrule
\multicolumn{5}{@{}l}{\textit{Semantic Encoder}}\\
ConvNeXt-T    & 68.13 & 46.87 & \textbf{33.98} & 64.29 \\
ConvNeXt-V2-Atto & 67.39 & 46.23 & 33.77 & 63.60 \\
\emph{none}   & 62.85 & 44.42 & 33.61 & 60.55 \\
\midrule
NAF           & 64.96 & 44.37 & 33.40 & 62.69 \\
\bottomrule
\end{tabular}
\caption[Ablations across four target-VFM families]{\textbf{Ablations across four target VFM
families.} Each row changes one component of the deployed configuration. All values are
mIoU$\uparrow$: Cityscapes linear probing on DINOv3-B and Franca-B, KITTI open-vocabulary transfer
on C-RADIO, KITTI DiffCut clustering on SSD-1B. \textbf{Bold} best per
column; the deployed row is shaded.}
\label{tab:gen-arena}
\end{table}

\FloatBarrier
\section{Qualitative Results}
\label{sec:qual}
\begin{figure*}[p]
\centering
\includegraphics[width=\textwidth]{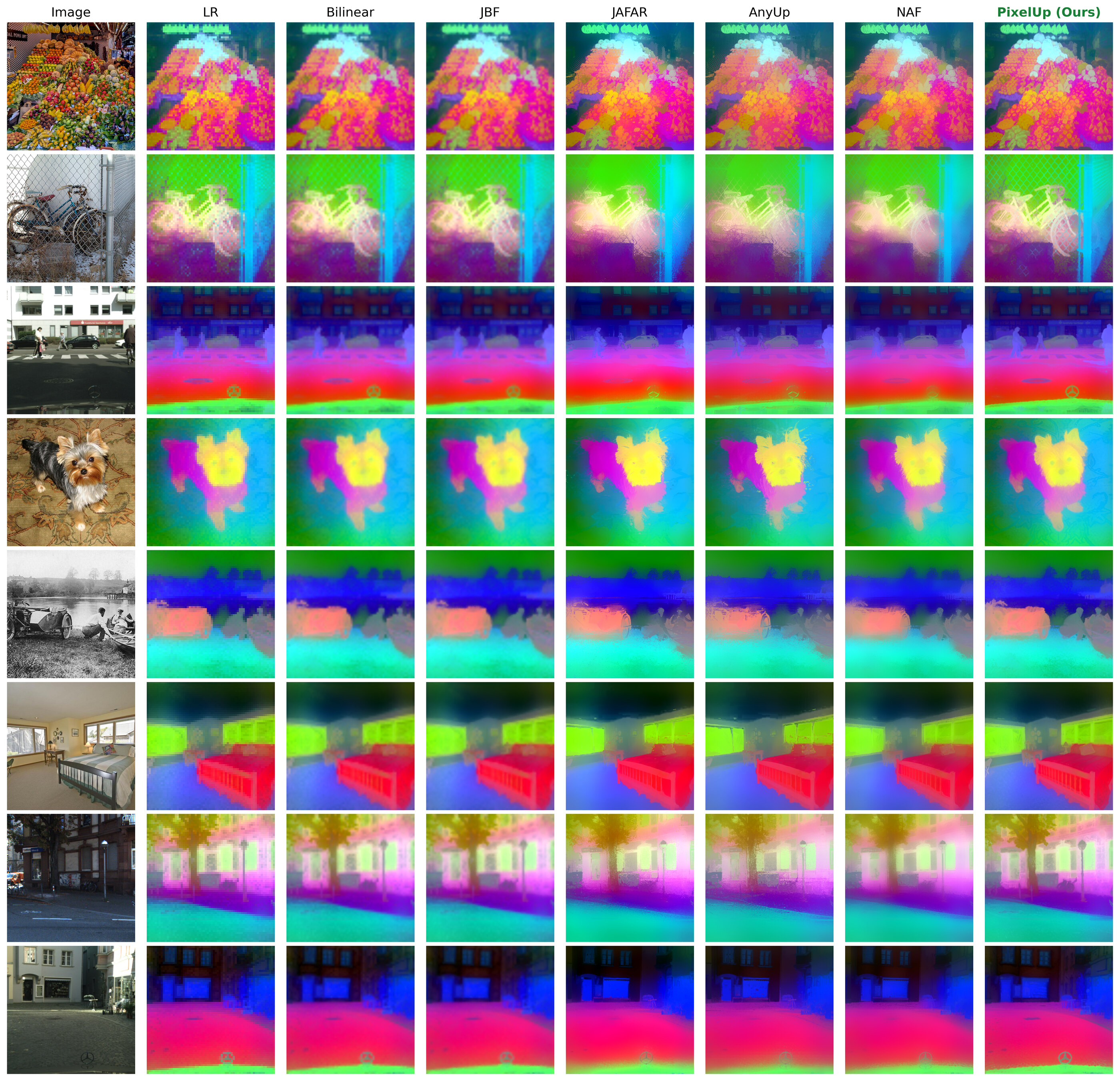}
\caption[Feature PCA across upsamplers]{\textbf{Upsampled features.} Each row is one image and
each column one upsampler, on a frozen DINOv3-L backbone. Each upsampled feature map is projected
onto its first three principal components and rendered as RGB; the PCA basis is shared across
methods within each row, so colors are comparable along a row.}
\label{fig:supp-pca}
\end{figure*}

\begin{figure*}[p]
\centering
\includegraphics[height=0.92\textheight]{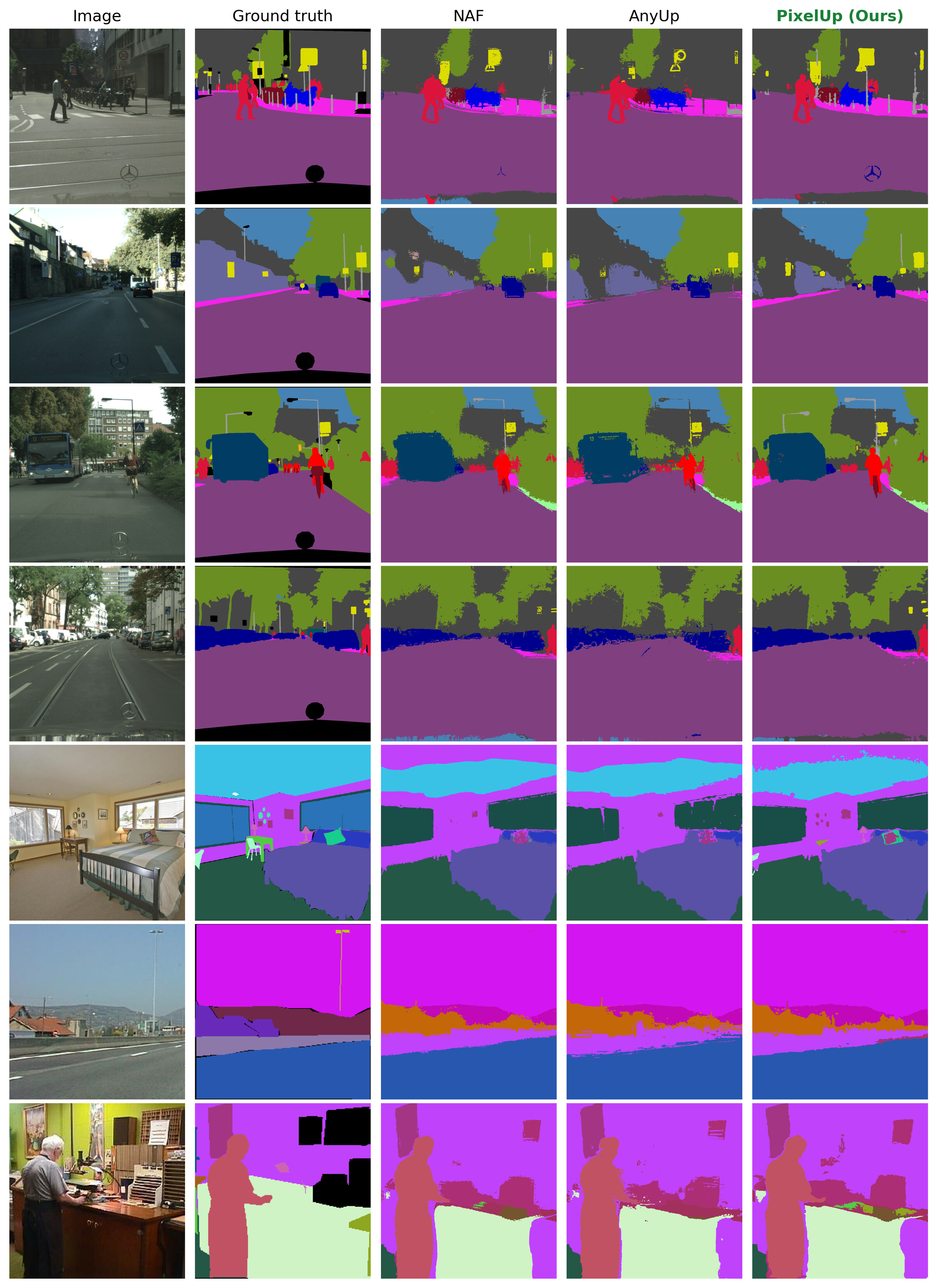}
\caption[Semantic segmentation predictions]{\textbf{Semantic segmentation.} Each row is one image
and each column one upsampler, shown beside the ground-truth label. Every column shares a frozen
DINOv3-B, each with its own trained probe.} 
\label{fig:supp-seg}
\end{figure*}

\begin{figure*}[p]
\centering
\includegraphics[height=0.92\textheight]{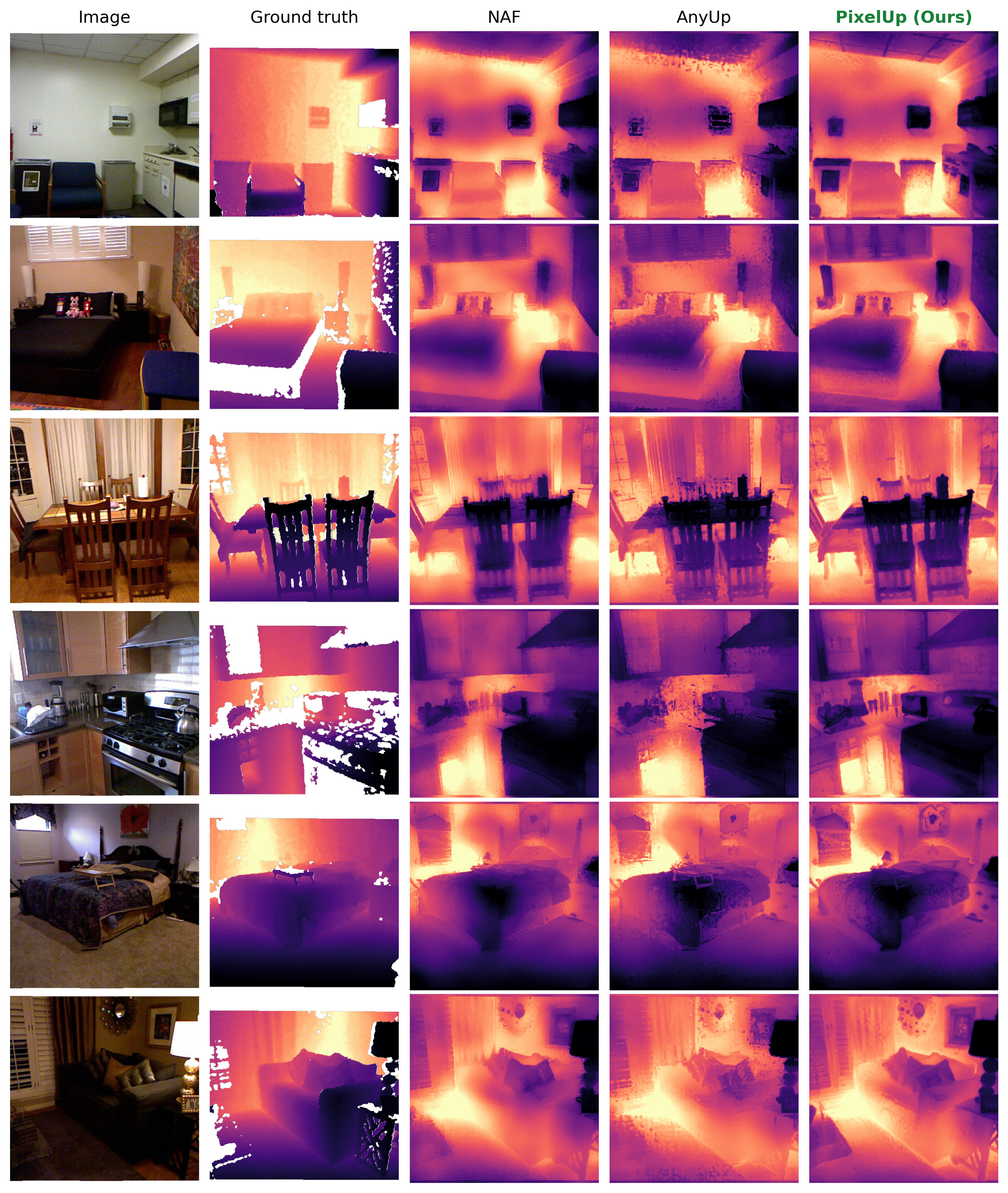}
\caption[Monocular depth predictions]{\textbf{Monocular depth.} NYUv2 test split; each row is one
image and each column one upsampler, shown beside the label. Every column shares a frozen
DINOv3-B, each with its own trained depth head.}
\label{fig:supp-depth}
\end{figure*}

\begin{figure*}[p]
\centering
\includegraphics[height=0.92\textheight]{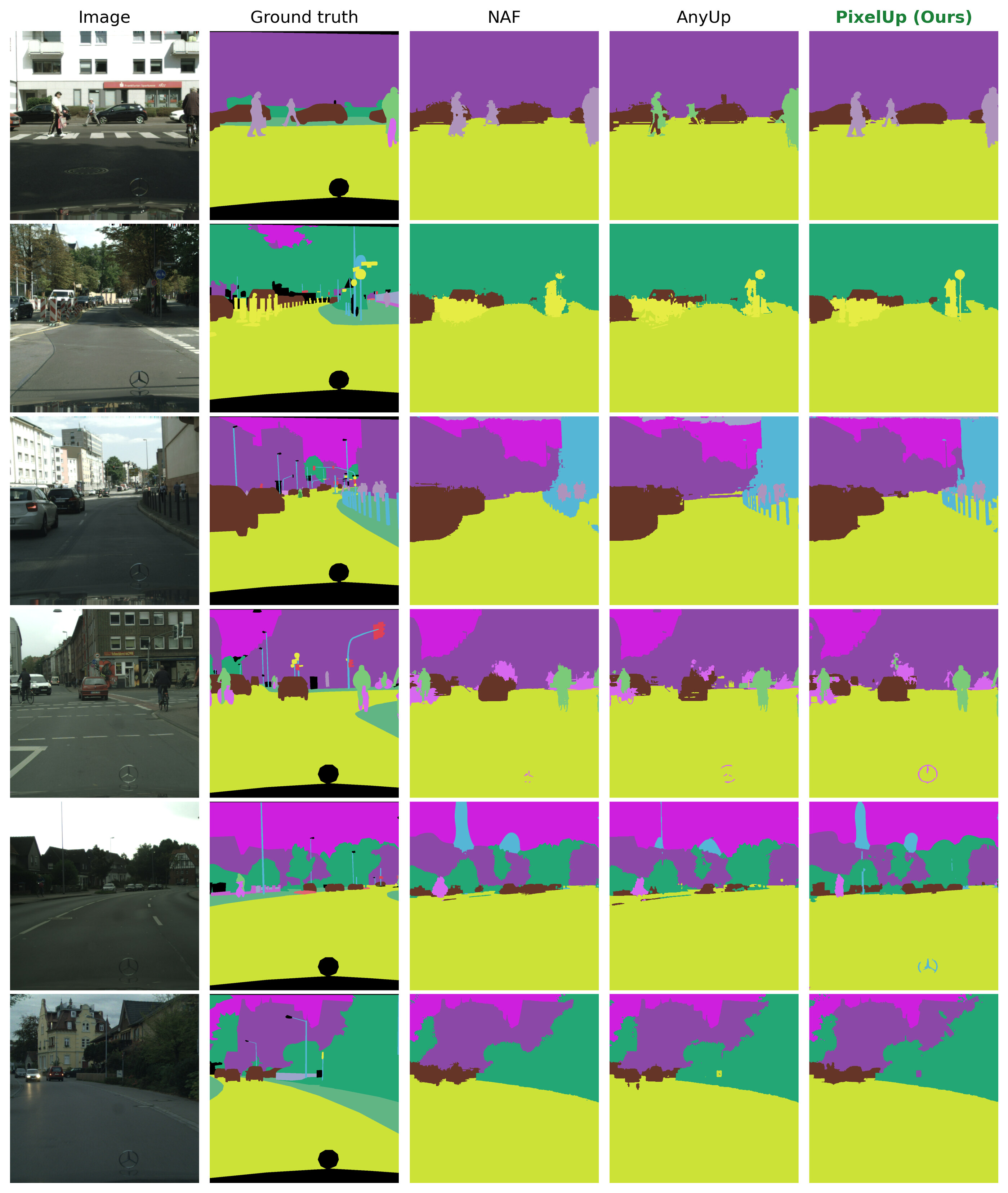}
\caption[Unsupervised segmentation]{\textbf{Unsupervised segmentation.} DiffCut on frozen SSD-1B
features, Cityscapes; each row is one image and each column one upsampler, and only the operation
that lifts the $32{\times}32$ features to the clustering grid changes.}
\label{fig:supp-unsup}
\end{figure*}

\begin{figure*}
\centering
\includegraphics[width=\textwidth]{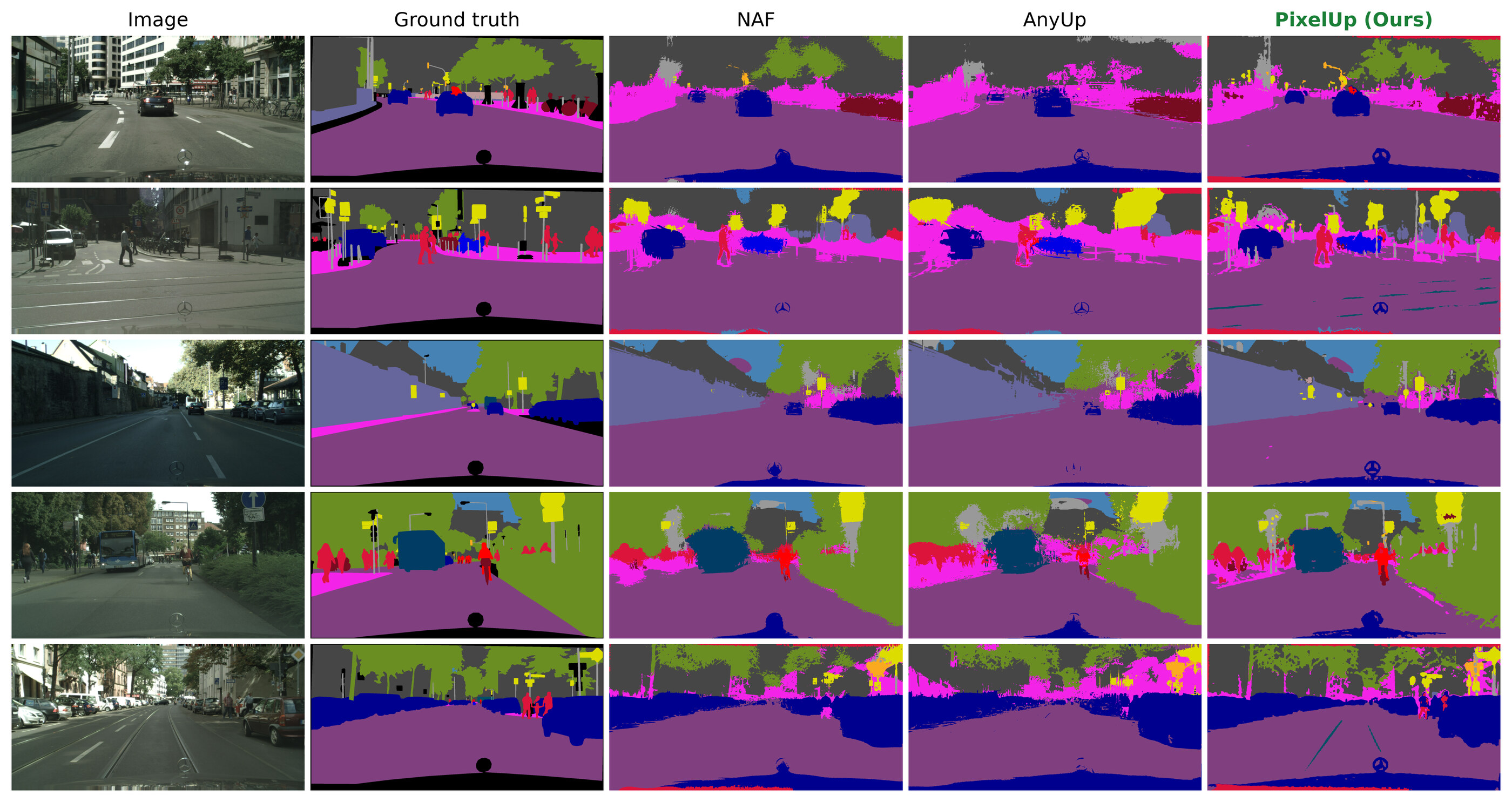}
\caption[Open-vocabulary segmentation]{\textbf{Open-vocabulary segmentation.} RADSeg with a frozen
C-RADIOv3-B encoder and its SigLIP2 adapter, scored against the Cityscapes label set; each row is
one image and each column one upsampler.}
\label{fig:supp-ov}
\end{figure*}

Figs.~\ref{fig:supp-pca}--\ref{fig:supp-ov} show the features and the predictions. Every
figure runs each method end-to-end under one shared configuration. Fig.~\ref{fig:supp-pca} compares the upsampled features themselves, with no task
head: inside an object \ours{} holds one near-uniform color, where the baselines reproduce image
texture and leak artifacts across object boundaries. The remaining figures show what that costs downstream. A $1{\times}1$ probe
cannot repair a feature error, so where a baseline's features bleed across a boundary the
prediction reproduces that bleed (Fig.~\ref{fig:supp-seg}), the depth discontinuity spreads over a
band several pixels wide (Fig.~\ref{fig:supp-depth}), a normalized cut breaks one surface into
several segments (Fig.~\ref{fig:supp-unsup}), and thin classes are absorbed into their
surroundings (Fig.~\ref{fig:supp-ov}). The classes to look at are the thin ones of
Tab.~\ref{tab:perclass}. These figures
illustrate the behavior the tables measure across different evaluations.

\bibliography{pixelup}